\documentclass{article}

\usepackage{microtype}
\usepackage{graphicx}
\usepackage{subcaption}
\usepackage{booktabs} 

\usepackage{hyperref}

\usepackage[preprint]{icml2026}

\usepackage{colortbl}
\usepackage{xcolor}
\usepackage{amsmath}
\usepackage{amssymb}
\usepackage{mathtools}
\usepackage{amsthm}
\usepackage{multirow}
\usepackage{rotating}
\usepackage{enumitem}
\usepackage{makecell}
\usepackage[capitalize,noabbrev]{cleveref}

\usepackage{listings}
\usepackage{xcolor}

\theoremstyle{plain}
\newtheorem{theorem}{Theorem}[section]
\newtheorem{proposition}[theorem]{Proposition}
\newtheorem{lemma}[theorem]{Lemma}

\theoremstyle{definition}

\theoremstyle{remark}

\usepackage[textsize=tiny]{todonotes}

\begin{document}

\twocolumn[
  \icmltitle{
  Differentiate-and-Inject:
Enhancing VLAs via Functional Differentiation Induced by In-Parameter Structural Reasoning
  }



  \icmlsetsymbol{equal}{*}

  \begin{icmlauthorlist}
    \icmlauthor{Jingyi Hou}{ustb}
    \icmlauthor{Leyu Zhou}{ustb}
    \icmlauthor{Chenchen Jing}{zjgy}
    \icmlauthor{Jinghan Yang}{ustb}
    \icmlauthor{Xinbo Yu}{ustb}
    \icmlauthor{Wei He}{ustb,bxk}
  \end{icmlauthorlist}

  \icmlaffiliation{ustb}{University of Science and Technology Beijing}
  \icmlaffiliation{zjgy}{Zhejiang University of Technology}
  \icmlaffiliation{bxk}{Beijing Information Science and Technology University}


  \icmlkeywords{Machine Learning, ICML}

  \vskip 0.3in
]



\printAffiliationsAndNotice{}  

\begin{abstract}
  As robots are expected to perform increasingly diverse tasks, they must understand not only low-level actions but also the higher-level structure that determines how a task should unfold.
  Existing vision-language-action (VLA) models struggle with this form of task-level reasoning.
  They either depend on prompt-based in-context decomposition, which is unstable and sensitive to linguistic variations, or end-to-end long-horizon training, which requires large-scale demonstrations and entangles task-level reasoning with low-level control.
  We present \underline{i}n-parameter \underline{S}tructured \underline{TA}sk \underline{R}easoning (iSTAR), a framework for enhancing VLA models via functional differentiation induced by in-parameter structural reasoning. 
  Instead of treating VLAs as monolithic policies, iSTAR embeds task-level semantic structure directly into model parameters, enabling differentiated task-level inference without external planners or handcrafted prompt inputs.
  This injected structure takes the form of implicit dynamic scene-graph knowledge that captures object relations, subtask semantics, and task-level dependencies in parameter space.
  Across diverse manipulation benchmarks, iSTAR achieves more reliable task decompositions and higher success rates than both in-context and end-to-end VLA baselines, demonstrating the effectiveness of parameter-space structural reasoning for functional differentiation and improved generalization across task variations.

\end{abstract}

\section{Introduction}

\begin{figure}[htbp]
  \centering
  \includegraphics[width=1.07\columnwidth, trim=50 25 0 25, clip]{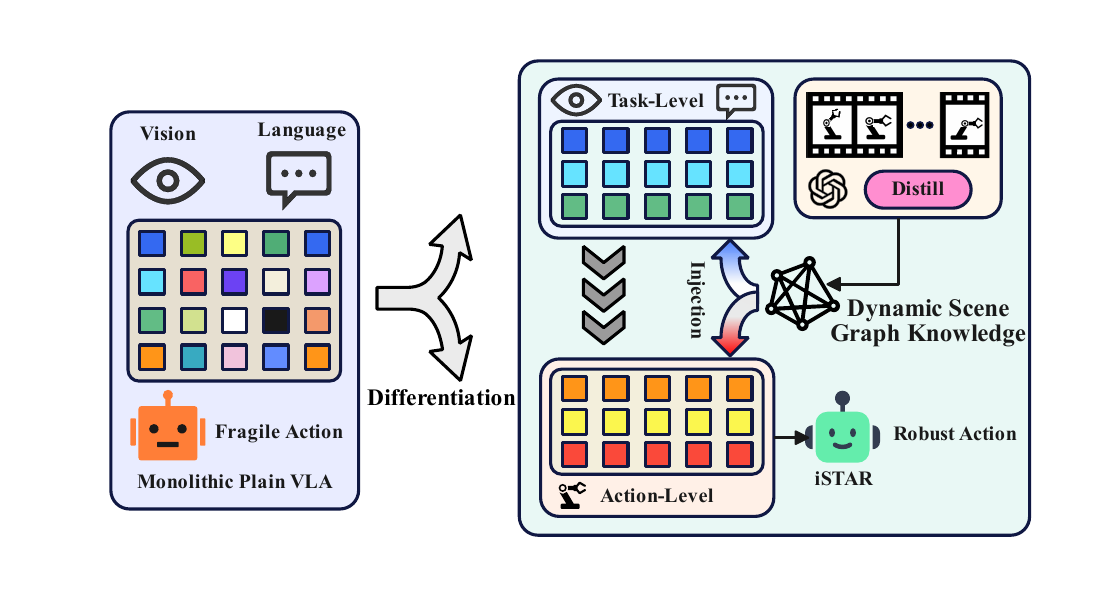}
  \caption{Our iSTAR transitions from a monolithic VLA with entangled reasoning to functionally differentiated modules where task-level structure (dynamic scene graphs) is injected into the model parameters.
  By the semantic-guided task resolution, iSTAR enhances the long-horizon reliability of VLA while maintaining end-to-end execution.}
  \vspace{-1em}
  \label{fig:figure1}
\end{figure}


Vision-language-action (VLA) models have emerged as a powerful paradigm for grounding natural language instructions in perception and control, enabling robots to perform a wide range of manipulation tasks~\cite{DBLP:journals/corr/abs-2210-03094,DBLP:conf/corl/KimPKXB0RFSVKBT24,shridhar2023perceiver}. 
Despite their success, most existing VLAs are used in a {monolithic} manner, where task-level reasoning and action execution are implicitly entangled within a single inference process. 
For long-horizon and procedurally constrained tasks, this coupling becomes a potential limitation, as errors in high-level reasoning can directly propagate into low-level control. 
Therefore, beyond simply scaling models or data, improving VLAs requires rethinking how their internal capabilities are organized.

Recent efforts to improve the reasoning ability of VLAs can be understood as introducing additional reasoning signals through in-context mechanisms. 
A common strategy enriches reasoning through language~\cite{ahn2022can,singh2023progprompt} or multimodal interaction~\cite{DBLP:journals/corr/abs-2502-19645,huang2025otter,intelligence2025pi_}, allowing models to reason about goals or task structure for action generation. 
While effective at guiding high-level intent, these reasoning processes rely primarily on prompt-based reasoning, rather than directly engaging with the action representations that govern physical execution.
As a result, task-level decisions are only indirectly reflected in action generation, making performance highly sensitive to how explicitly the instruction specifies task structure.
Other methods~\cite{zhao2025cot,10658623} perform reasoning in visual spaces to model object relations and temporal structure more explicitly. 
Although visual representations can faithfully capture physical scenes, reasoning within such high-dimensional manifolds is computationally demanding, making it less efficient to distill and manipulate the task-relevant structure needed for action execution. 
Consequently, there is a lack of a reusable mechanism for resolving and reusing task-level semantic structure at the level where actions are actually generated.

This perspective leads to the question of how a monolithic VLA can be enhanced to explicitly resolve task-level structure while preserving end-to-end action generation.
A closer look reveals that, although modern VLAs inherit strong reasoning capabilities from their vision–language components, such capabilities are often underutilized when task understanding and action execution are entangled within a single inference process.
In this work, we argue that enhancing VLAs requires a different design principle—\emph{functional differentiation induced by in-parameter structural reasoning}, as illustrated in Figure~\ref{fig:figure1}.
We differentiate the functional roles of a single VLA by embedding task-level semantic structure directly into its parameters without introducing external planners or separating reasoning and control into distinct models.
This enables task-level reasoning to be resolved explicitly and reused during inference, while preserving the expressive capacity of end-to-end VLA models.

Guided by this principle, we introduce \underline{i}n-parameter \underline{S}tructured \underline{TA}sk \underline{R}easoning (iSTAR), a framework that realizes functional differentiation through in-parameter structural reasoning.
iSTAR injects implicit and dynamic scene-graph knowledge into model parameters, capturing object-centric relations, subtask semantics, and task-phase dependencies in a form that can be reused across task instances.
This structural knowledge is distilled from vision-language models (VLMs) that observes task execution videos and produces subtask-level semantic descriptions, which are then used to guide the learning of dynamic scene structure within the VLA.
By internalizing such distilled structure, iSTAR resolves task-level reasoning to improve the reliability and generalization of long-horizon behavior without relying on handcrafted prompts and external planners.
Experiments across diverse manipulation benchmarks demonstrate that iSTAR consistently outperforms prompt-based and end-to-end VLA baselines, highlighting in-parameter structural reasoning as a principled mechanism for enhancing VLA models. 
\footnote{Our code is available at \url{https://github.com/hugezhuwo/iSTAR_code}.}

\vspace{-1em}

\section{Related Work}


Robot manipulation planning has traditionally been studied under hierarchical task and motion planning (TAMP), which decomposes long-horizon tasks into symbolic task-level decisions and continuous motion generation~\cite{kaelbling2011hierarchical}. 
Learning-based extensions incorporate visual perception and neural representations to improve robustness and scalability~\cite{zhu2021hierarchical}, but typically rely on predefined task structure.
With the emergence of LLMs, recent work explores language-driven planning for robotic manipulation. 
Methods such as SayCan~\cite{ahn2022can} and ProgPrompt~\cite{singh2023progprompt} generate high-level action sequences from natural language, while related approaches translate instructions into symbolic plans~\cite{zhou2024isr,guo2024castl}. 
In these methods, task-level reasoning is generally externalized and executed separately from action generation.




More recently, end-to-end VLA models~\cite{DBLP:journals/corr/abs-2210-03094,DBLP:conf/corl/KimPKXB0RFSVKBT24,intelligence2025pi_,shridhar2023perceiver,11128272} encode planning implicitly within action prediction, with recent work further studying post-training or online improvement of such policies~\cite{DBLP:journals/corr/abs-2506-15799}. 
While effective at scale, this monolithic formulation tightly entangles task-level reasoning with low-level control in latent action spaces, making it difficult to isolate and reuse task-level structure across task compositions and long-horizon variations.
Hierarchical VLA approaches introduce intermediate representations to decompose planning and execution~\cite{DBLP:journals/corr/abs-2502-19417,DBLP:conf/iclr/0038DZJMG0F00025,dalal2024plan,chen2025robohorizon,zhao2025cot,10658623}. 
These methods relieve high-level models from fine-grained control, but their intermediate representations are usually limited to paths, poses, or visual subgoals, which makes it difficult to capture compositional semantic structure. 
Related work further explores efficient control via architectural~\cite{DBLP:journals/corr/abs-2503-20384} or modality augmentation~\cite{DBLP:journals/corr/abs-2505-22159}. 
OpenVLA-OFT \cite{DBLP:journals/corr/abs-2502-19645} improves task performance by parameter-efficiently fine-tuning a VLA with visual reasoning \cite{DBLP:conf/aaai/PerezSVDC18}.
By explicitly resolving and reusing task-level semantic structure \emph{in parameter space}, iSTAR provides a principled pathway toward systematic compositional generalization in long-horizon vision-language-action tasks.

\begin{figure*}[htbp]
  \centering
  \includegraphics[width=2\columnwidth, trim=0 40 0 0, clip]{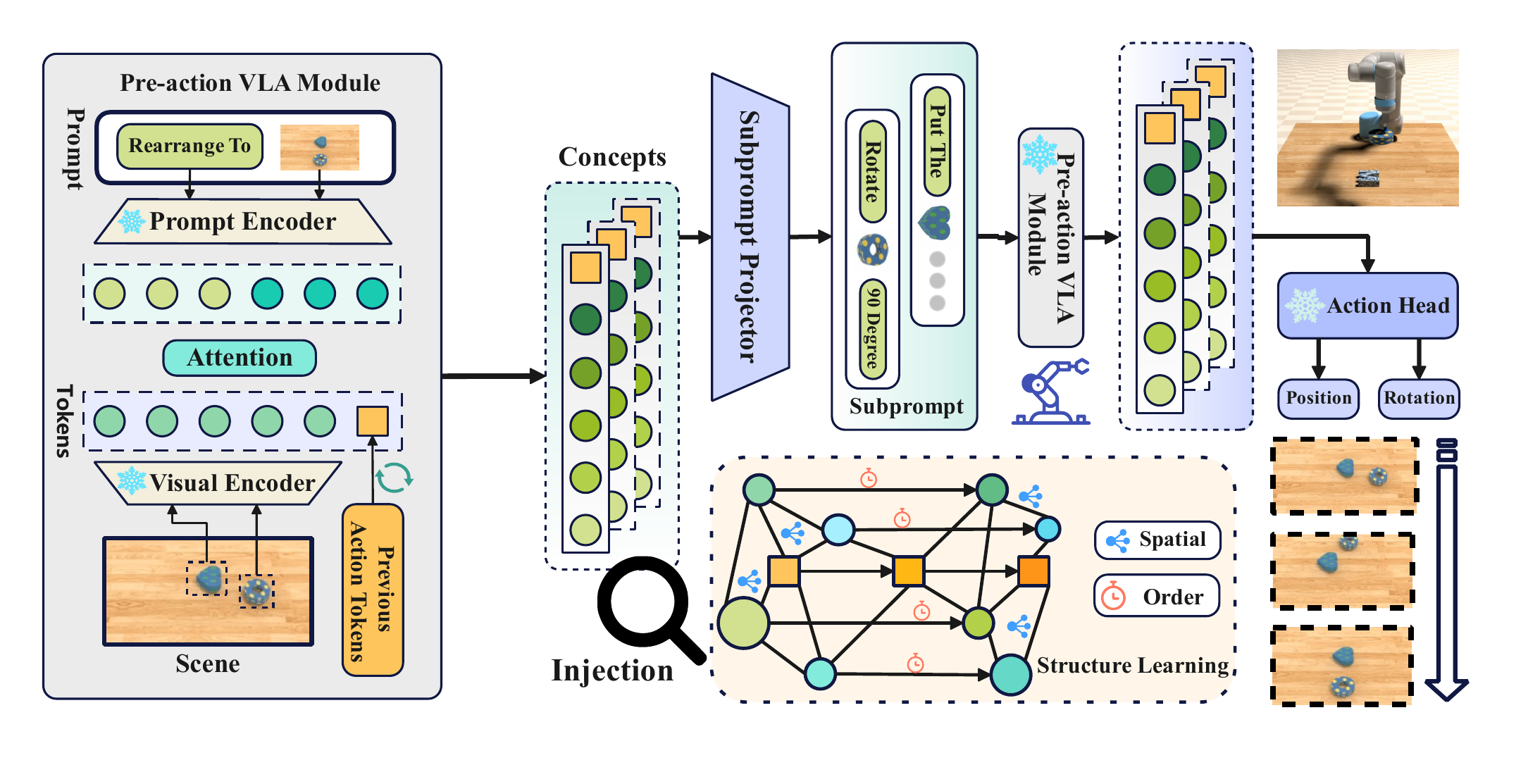}
  \caption{Overview of the proposed iSTAR framework. 
  }
  \label{fig:istar_overview}
\end{figure*}


\section{Method}
\label{sec:method}


Motivated by the limitations of monolithic VLAs discussed above, the core goal of iSTAR is to explicitly resolve task-level semantic structure before action execution, while keeping such reasoning internal to the model parameters and compatible with end-to-end VLA inference.
To this end, iSTAR introduces functional differentiation within a pretrained VLA by assigning distinct roles to different instances of the same backbone.
Specifically, one instance operates on the VLA backbone before the action head (the \emph{pre-action VLA module}) to perform semantic concept reasoning, while another instance is used for action generation conditioned on the resolved task structure.
Figure~\ref{fig:istar_overview} illustrates an overview of this differentiated inference structure. 

Given a multimodal instruction $c$ and visual observations $I$, iSTAR first extracts object- and action-centric concept representations from the pre-action VLA module.
To guide these representations toward task-relevant semantic reasoning, we inject dynamic concept-graph structure into this module, enabling structured reasoning among concepts and enriching concept representations that capture the semantics required for the current task stage.
These structure-aware concept representations are then mapped to subtask-level textual descriptions (subprompts), which serve as explicit semantic commitments.
Conditioned on these subtask descriptions as prompt embeddings, the action-generating VLA solves a simplified short-horizon decision problem.
Overall, iSTAR decomposes a monolithic VLA inference process into differentiated concept reasoning and action execution stages, while keeping both stages within the same VLA parameter family and without introducing external planners.

\subsection{Concept Extraction}
\label{sec:concept_extraction}

Our iSTAR uses the pre-action VLA module to perform semantic concept reasoning.
This module is responsible for extracting object- and action-centric representations that summarize the task-relevant semantics at each stage.
Given a task instruction $c$ and a sequence of visual observations $I=\{I_t\}_{t=1}^T$, we maintain at each timestep $t$ an object embedding $o_t$ derived from the current observation and an action embedding $a_{t-1}$ carried over from the previous step.
These embeddings are processed by a multimodal autoregressive encoder, i.e., the pre-action VLA module, producing object- and action-centric concept embeddings
\begin{equation}
(v_t^{\mathrm{obj}}, v_t^{\mathrm{act}})
= \mathrm{Enc}\big(c,\, o_t,\, a_{t-1}\big),
\end{equation}
where $a_{t-1}=0$ for $t=1$ and equals the previous action concept $v_{t-1}^{\mathrm{act}}$ for $t>1$.
The resulting $v_t^{\mathrm{obj}}$ remains an object-centric representation grounded in the current observation, while $v_t^{\mathrm{act}}$ encodes a high-level action intent inferred at timestep $t$.
The autoregressive loop is closed by feeding $v_t^{\mathrm{act}}$ forward as $a_t$ for the next step, allowing concept representations to evolve coherently across timesteps.
Collecting all concept embeddings yields the node set
\[
\mathcal{V} = \{v_t^{\mathrm{obj}}, v_t^{\mathrm{act}} \mid t = 1,\dots,T\},
\]
which constitutes the semantic concept set over which iSTAR performs in-parameter structural reasoning.
This set serves as the input to the dynamic implicit concept graph module described in \cref{sec:graph}.
For clarity, unless otherwise specified, we assume all concept representations correspond to the current timestep, and the explicit time index $t$ is omitted in the following discussion.

\subsection{Dynamic Implicit Concept Graph Construction}
\label{sec:graph}

Given the concept set $\mathcal{V} = \{v_1,\dots,v_N\}$, the goal of the implicit dynamic graph module is to infer the task-level structure that determines which objects and action cues are relevant for the current subtask, as well as how these concepts interact temporally and relationally. 

\paragraph{Attribute gating.}
To select concepts relevant to the current subtask, each node is modulated by an attribute gate:
\begin{equation}
\tilde v_i = v_i \odot \sigma(W_{\mathrm{attr}} v_i).
\label{eq:attr-gate}
\end{equation}
This operation suppresses dimensions unrelated to the underlying task semantics and highlights
candidate objects and action cues.

\paragraph{Dynamic positional encoding.}
Temporal structure across subtasks is modeled through a recurrent state that accumulates
information across decoding steps:
\begin{equation}
h_{t,i},\; v_{i}^{(\mathrm{pe})}
= \mathrm{GRU}\!\big(h_{t-1,i},\, \tilde v_i\big),
\label{eq:dyn-pe}
\end{equation}
Unlike static positional encodings, these representations evolve with the inferred subtask sequence and capture temporal ordering cues.

\paragraph{Order gating and fusion.}
To refine temporal relevance, a second gate modulates the positional representation:
\begin{equation}
\hat v_i^{(\mathrm{pe})}
= v_i^{(\mathrm{pe})} \odot \sigma(W_{\mathrm{order}} v_i^{(\mathrm{pe})}).
\label{eq:order-gate}
\end{equation}
The gated representations are then fused with the attribute-filtered nodes:
\begin{equation}
v_i^{(\mathrm{fused})}
= \tilde v_i + \hat v_i^{(\mathrm{pe})}.
\label{eq:fused}
\end{equation}
The resulting node states encode both object relevance and temporal phase information.

\paragraph{Implicit relational reasoning.}
Relational structure is captured through multi-head attention, optionally incorporating edge-level
representations when available:
\begin{equation}
\begin{aligned}
&V^{(\mathrm{out})}
= \\
&\mathrm{Attn}\!\left(
V^{(\mathrm{fused})},\;
V^{(\mathrm{fused})} + E^{(\mathrm{fused})},\;
V^{(\mathrm{fused})} + E^{(\mathrm{fused})}
\right),
\end{aligned}
\label{eq:attn-general}
\end{equation}
where $E^{(\mathrm{fused})}$ denotes optional edge embeddings (if provided).  
The resulting attention patterns implement latent message passing within the implicit concept graph.

\paragraph{Structure-learning objective.}
Let $p_t$ denote the subtask description at decoding step $t$, represented either as a textual
embedding or a multimodal prompt embedding depending on the downstream VLA interface. To guide the
implicit graph toward task-relevant structure, we introduce an auxiliary objective that aligns
concept relevance with subtask semantics while maintaining uncertainty over temporal ordering:
\begin{equation}
\begin{aligned}
&\mathcal{L}_{\mathrm{struct}}
= \\
& - \log
\frac{\exp(\mathrm{sim}(\tilde v_i,\, p_t))}
     {\sum_j \exp(\mathrm{sim}(\tilde v_j,\, p_t))}
\;-\;
\lambda\, \mathbb{H}\!\left(
\sigma(W_{\mathrm{order}} v_i^{(\mathrm{pe})})
\right).
\end{aligned}
\label{eq:struct-loss}
\end{equation}
The first term encourages the gated concept representations $\tilde v_i$ to align with the semantic
content of the current subtask, promoting correct object and action selection. The second term
regularizes the entropy of the order gate, preventing premature commitment to a fixed temporal
structure and allowing the model to explore multiple plausible subtask orderings during learning.

\subsection{Subtask Prompt Projector}
\label{sec:projector}

After dynamic structural reasoning, the fused concept representations
$V^{(\mathrm{fused})}=\{v^{(\mathrm{fused})}_1,\dots,v^{(\mathrm{fused})}_N\}$
encode object relevance, temporal phase, and relational context for the current subtask.
The subtask prompt projector maps these structure-aware semantic representations to 
subtask-level language space that serves as semantic commitment for downstream action
generation.

In fact, the projector acts more like a
\emph{semantic embedding mapper}, translating representations from the semantic concept space
into the language embedding space expected by the downstream VLA.
Since subtask prompts correspond to short, structurally constrained intents, the projection can be
implemented in either a non-autoregressive or autoregressive manner in embedding space, without
encountering the challenges of long-form language modeling.
The resulting subtask embedding 
is used to condition the action-generating VLA, reducing the
burden of long-horizon semantic reasoning during action decoding.

\subsection{Subtask Prompt Distillation and Supervision}

Most VLA datasets do not provide explicit subtask-level prompts, but they do include step-wise visual observations aligned with action trajectories.
We leverage this property to construct subtask prompt supervision via online language model distillation without additional manual annotation.
For each subtask step, we collect several textual descriptions from a powerful VLM of the task context, which is used as supervision for training the subtask prompt decoder.
This enables the decoder to learn a compact subtask-level description aligned with the underlying execution, while remaining agnostic to the full multimodal input.

For VIMA-Bench, where prompts are inherently multimodal, we first obtain descriptions with placeholders and then ground them using the provided action trajectories and object segmentation annotations.
Object masks and spatial references are recovered from the dataset annotations, allowing reconstruction of multimodal subtask prompts compatible with token-based VLAs such as VIMA.

\section{Theory and Extensions}
\label{sec:theory}

The empirical gains observed in real-world tasks suggest that explicitly modeling semantic commitment before action decoding improves generalization.
In this section, we formalize this intuition and analyze why structured semantic reasoning yields advantages in large-horizon and low-information regimes.


\subsection{Structured Advantage in the Large-Horizon Regime}

We consider a structured policy $\hat\pi=(\hat f,\hat r,\hat h)$, which first predicts a semantic graph $\hat G_t$, realizes it into a subprompt $\hat p_t$, and then decodes a one-step action
$a_t \sim \hat h(\cdot \mid I_t,\hat p_t)$.
We compare it to an end-to-end policy $\hat\pi_{\mathrm{e2e}}$ that directly predicts discrete action tokens $a_t \in \mathcal A$.

Assume the expert policy admits latent semantic witnesses $(G_t^*,p_t^*)$ such that the expert action distribution satisfies
$a_t^* \sim h^*(\cdot \mid I_t,p_t^*)$,
where $p_t^* = r^*(G_t^*)$.
Under standard performance sensitivity arguments, according to standard analyses in imitation learning and structured prediction \cite{pmlr-v9-ross10a,ross2011reduction},
the performance gap of the structured policy admits the decomposition
\begin{equation}
J(\pi^*) - J(\hat\pi)
\;\le\;
C_a\,\epsilon_a
\;+\;
C_r\,\epsilon_r
\;+\;
C_G\,\epsilon_G,
\label{eq:error_decomposition}
\end{equation}
where $\epsilon_a$ denotes the one-step action decoding error under the correct semantic commitment,
$\epsilon_r$ denotes the realization error from semantic graph to subprompt,
and $\epsilon_G$ denotes the semantic graph prediction error.

Conditioned on the correct semantic commitment, action decoding reduces to a one-step supervised learning problem, so $\epsilon_a$ does not require modeling long-horizon dependencies.
If the realization step is implemented via constrained templates or grammar, $\epsilon_r$ is bounded and can be treated as a lower-order term.
As a result, the dominant contribution arises from $\epsilon_G$.

\begin{proposition}[Structured advantage for large horizons]
\label{prop:structured_advantage}
Consider discrete action tokens $a_t \in \mathcal A$.
Suppose the effective complexity of semantic commitment over a horizon $T$ scales as
\begin{equation}
\log \mathcal N(\mathcal F_G,\varepsilon)
=
\Theta\!\left(T\log|\mathcal C|\right),
\end{equation}
while an end-to-end policy predicting action tokens induces
\begin{equation}
\log \mathcal N(\Pi_{\mathrm{e2e}},\varepsilon)
=
\Theta\!\left(T\log|\mathcal A|\right).
\end{equation}
If $\log|\mathcal C| < \log|\mathcal A|$ and the supervision budgets are of the same order, then there exists $T_0$ such that for all $T \ge T_0$, the generalization bound of the structured policy is strictly tighter than that of the end-to-end policy.
\end{proposition}

\paragraph{Proof Sketch.}
We outline the main steps of the comparison; full proofs are provided in Appendix~A.

We begin by bounding the structured policy.
Using the decomposition in Eq.~\eqref{eq:error_decomposition}, the global performance gap is reduced to the sum of three terms corresponding to action decoding, realization, and semantic graph prediction.
The first two terms are well-conditioned: conditioned on the correct semantic commitment $p_t^*$, the action decoder only needs to model a one-step conditional distribution $(I_t,p_t^*) \mapsto a_t^*$, and therefore its generalization behavior does not scale with the task horizon.
Similarly, when the realization from graph to subprompt is implemented via templates or constrained grammar, the induced realization error remains bounded and does not accumulate over time.

The remaining term, $\epsilon_G$, captures the difficulty of predicting semantic commitments.
Applying standard uniform convergence arguments to the semantic graph predictor yields a bound of the form
\begin{equation}
\epsilon_G
\;\le\;
\widehat{\epsilon}_G
+
O\!\left(
\sqrt{
\frac{T\log|\mathcal C|}{n_G}
}
\right),
\end{equation}
where $\widehat{\epsilon}_G$ is the empirical graph prediction error and $n_G$ is the number of supervised graph samples.
Substituting this bound into Eq.~\eqref{eq:error_decomposition} and absorbing horizon-independent terms gives an overall structured bound whose leading term scales as $\sqrt{T\log|\mathcal C|}$.

For the end-to-end policy, directly predicting a length-$T$ sequence of discrete action tokens induces an effective output space of size $|\mathcal A|^T$.
As a result, standard complexity-based generalization arguments yield a bound whose leading term scales as $\sqrt{T\log|\mathcal A|}$.

Since $\log|\mathcal C| < \log|\mathcal A|$, the dominant term of the structured bound grows strictly more slowly than that of the end-to-end bound.
Moreover, as analyzed above, the remaining two error components of the structured policy ($\epsilon_r$ and $\epsilon_a$) are horizon-insensitive and therefore do not accumulate as the task horizon increases.
Consequently, there exists a horizon length $T_0$ such that for all $T \ge T_0$, the structured policy admits a strictly tighter generalization bound than the end-to-end policy.
\hfill$\square$


To complete the argument in Proposition~\ref{prop:structured_advantage}, we briefly explain why the condition $\log|\mathcal C| < \log|\mathcal A|$ holds in practice.
The semantic commitment space $\mathcal C$ is constructed from a set of discrete semantic attributes,
\[
\mathcal C = \{(c_1,\dots,c_K)\mid c_k\in\mathcal C_k\}.
\]
Crucially, a semantic commitment is specified by selecting one value for each attribute, so $\mathcal C$ scales as
\[
|\mathcal C|
\;=\;
\sum_{k=1}^K |\mathcal C_k|
\ll \;
\!\prod_{k=1}^K |\mathcal C_k|.
\;
\]
This reflects that semantic graphs encode abstract attributes (e.g., object identity, relations, constrains) whose contributions accumulate additively.
The multiplicative space $\prod_k |\mathcal C_k|$ can be viewed as a discrete combinatorial abstraction of executable behaviors, but it is already a simplification of the true action space $\mathcal A$.
In practice, $\mathcal A$ represents low-level physical control and is typically a discretization of a high-dimensional continuous space.
Its effective complexity therefore grows with both control dimensionality and resolution, far exceeding that of the semantic space.
Consequently, the complexity of the semantic commitment space is much smaller than that of the action space, justifying the regime
\[
\log|\mathcal C| \ll \log|\mathcal A|
\]
assumed in Proposition~\ref{prop:structured_advantage}.

\subsection{Advantage in the Low-Information  Regime}


Beyond long-horizon execution, structured decomposition is also advantageous in tasks with low perceptual information but high semantic reasoning demands, such as relational instructions (e.g., “place all objects with the same shape as X into the bowl”) or procedural tasks (e.g., making a cup of coffee).
In such regimes, the challenge lies not in action execution but in resolving semantic ambiguity under weak supervision. 
This difficulty has been widely observed in sequence-to-sequence models, which often succeed at surface-level generalization but fail when correct behavior requires extracting and systematically applying abstract compositional rules rather than exploiting latent statistical correlations (e.g., \citeauthor{DBLP:conf/icml/LakeB18}, \citeyear{DBLP:conf/icml/LakeB18}).
Our method resolves this ambiguity at the semantic level, enabling systematic reuse of abstract rules and yielding improved compositional generalization under limited perceptual information.

\subsection{Extending iSTAR beyond VIMA}
\label{sec:extension}

Although Section~\ref{sec:method} uses VIMA as an example, iSTAR is architecture-agnostic and can be applied to a
broad range of VLA backbones.
The only requirement is access to intermediate representations before the action head, from which
task-relevant semantic information can be extracted.
VIMA is used for illustration due to its explicit multimodal token structure and the suitability of
VIMA-Bench for evaluating task-level reasoning.

For decoder-based VLAs, iSTAR follows the same principle by operating on backbone representations
before the action head to produce structure-aware subtask embeddings.
For large-scale VLAs, subtask embeddings can be extracted from higher-level hidden states close to the
action head, avoiding excessive reuse of lower-level features and unnecessary parameter growth.








\section{Experiments}

\subsection{Experimental Overview}

We evaluate our method across two public benchmarks and a real-world robotic system, with each setting targeting a complementary aspect of vision-language-action reasoning.
VIMA-Bench and LIBERO are designed to assess visual reasoning and perception-conditioned manipulation under diverse task and object configurations.
These benchmarks emphasize grounding language instructions in visual scenes, handling object relations, spatial arrangements, and long-horizon execution under standardized task formulations, enabling controlled comparisons with prior work.

While benchmark evaluations focus on visual grounding and execution under fixed task specifications, they provide limited coverage of task-level generalization when execution structure must be inferred beyond the given instruction.
To complement this, we further conduct real-world robotic experiments that explicitly probe task-level reasoning and generalization.
In real-world settings, prompts specify only the desired goal state and omit prerequisite subtasks or execution order, requiring the model to infer execution structure from the current scene.
Moreover, real-world execution involves irreversible errors and environmental variability, making it a necessary testbed for evaluating whether task-level semantic structure learned in parameter space can generalize beyond benchmark distributions.

Together, these experimental settings provide a comprehensive evaluation of visual reasoning, action grounding, and task-level generalization, allowing us to assess the practical effectiveness of in-parameter structural reasoning from simulation to real-world deployment.

\subsection{Experiments on VIMA-Bench}

Experiments on VIMA-Bench organize tasks into 4 levels of increasing
difficulty (L1--L4), reflecting progressively more complex compositional structure and longer horizons, following the official task stratification provided by VIMA-Bench.

To ensure fair comparison, all VIMA-based results reported in this paper are obtained by fine-tuning the official pretrained VIMA models on the task configurations used in our experiments.
Detailed task settings and per-task success rates are provided in
Appendix~\ref{app:extend_vima}.
VIMA provides two backbone configurations with 2M and 4M model sizes, denoted as VIMA (Small) and VIMA (Large), respectively.
In our experiments, iSTAR is built upon the VIMA (Small) backbone.
Due to the additional structural reasoning components introduced by iSTAR, the total parameter count of iSTAR is comparable to that of the VIMA (Large) model.
We therefore report results for both VIMA (Small) and VIMA (Large) as reference baselines.

\paragraph{iSTAR variants for ablation.}
All iSTAR ablation variants share the same backbone and differ only in the components of the dynamic implicit concept graph.
Specifically, we evaluate: 






\textbf{iSTAR (Stacked VIMA).} 
This variant tests whether performance gains arise from increased model depth alone, by stacking a pre-action VIMA with a frozen pretrained VIMA, without subtask supervision or structural reasoning.

\textbf{iSTAR (w/o Graph).}
This variant removes the dynamic implicit graph and all structural losses, while retaining subtask embeddings to condition the action VLA, isolating the effect of subtask-level supervision without explicit structure.

\textbf{iSTAR (w/o Concept Alignment).}
This variant retains the implicit graph but removes the concept–subtask alignment term, eliminating explicit supervision of concept relevance.

 \textbf{iSTAR (w/o Order Entropy).}
This variant retains the implicit graph but removes order entropy regularization, allowing early commitment to a fixed subtask ordering.

\begin{table}[t]
\centering
\caption{Success rate (\%) across VIMA-Bench task levels (L1--L4). Best results are in \textbf{bold}, second-best are \underline{underlined}.}
\label{tab:l1_l4}
\setlength{\tabcolsep}{6pt}
\small
\renewcommand{\arraystretch}{1.15}

\begin{tabular}{lcccc}
\toprule
\textbf{Method} & \textbf{L1} & \textbf{L2} & \textbf{L3} & \textbf{L4} \\
\midrule
VIMA {\scriptsize (Small)}                        & 72.4 & 73.8 & 71.1 & 35.0 \\
VIMA {(\scriptsize Large)}                        & 73.9 & 74.5 & 72.0 & \textbf{59.0} \\
\midrule
iSTAR {\scriptsize (Stacked VIMA)}                & 72.4 & 73.6 & 72.4 & 34.0 \\
iSTAR {\scriptsize (w/o Graph)}                   & 74.8 & 76.0 & 74.9 & 37.7 \\
iSTAR {\scriptsize (w/o Concept Alignment)}       & \underline{77.8} & 77.5 & \underline{77.0} & 40.0 \\
iSTAR {\scriptsize (w/o Order Entropy) }          & 76.4 &\underline{ 77.6} & 76.3 & 42.7 \\
iSTAR {\scriptsize (Ours)}                        & \textbf{78.6} & \textbf{79.2} & \textbf{78.3} & \underline{44.3} \\
\bottomrule
\end{tabular}
\end{table}

\paragraph{Result Analysis.}
The ablation results in Table~\ref{tab:l1_l4} highlight the respective roles of intermediate supervision and structural inductive biases in iSTAR.
Simply stacking two VIMA backbones without intermediate supervision yields performance comparable to VIMA (Large) at L1-L3 level, indicating that increased model depth alone provides limited benefits.
iSTAR (w/o Graph) already leads to noticeable improvements, suggesting that subtask-level supervision provides a strong semantic signal that the VLA backbone can partially exploit.
iSTAR (w/o Concept Alignment) and iSTAR (w/o Order Entropy) show higher performance than iSTAR (w/o Graph).
While the averaged metrics aggregate tasks of varying difficulty within each level, we observe consistent and non-trivial improvements across all task levels.
In particular, the gains become more obvious at higher task levels that involve more objects and longer execution horizons.
These results indicate that explicitly supervising concept relations and maintaining uncertainty over subtask ordering are especially effective in complex, long-horizon settings.
Overall, the full iSTAR model achieves the best overall performance.


We further observe that VIMA (Large) exhibits a notably stronger performance on L4.
A closer inspection reveals that one task (Task 08, ``novel\_adj\_and\_noun'' described in \citet{DBLP:journals/corr/abs-2210-03094}) of L4 contains prompts requiring the most complex reasoning over all tasks,
such as one-shot concept grounding and attribute comparison, which place heavy demands on the
vision-language encoding capacity rather than on action generation.
This suggests that the observed performance gap is primarily caused by insufficient visual-language encoding capacity, which leads to underfitting on tasks requiring complex multimodal reasoning, rather than deficiencies in task-level reasoning itself.
iSTAR is designed to explicitly elicit and organize the reasoning capabilities already present in
the underlying VLA, but cannot compensate for missing representational capacity; as a result, while
it substantially improves performance over smaller backbones, its effectiveness is ultimately
bounded by the model’s intrinsic visual-language reasoning capacity.

Consistent with this hypothesis, in a supplementary experiment we replace the task-level pre-action VLA with a VIMA large backbone while keeping the rest of iSTAR unchanged (consistent with the 2M backbone), which improves the success rate from 85\% to 97\%.

\subsection{Experiments on LIBERO}

Table~\ref{tab:sr_libero} reports success rates on the LIBERO benchmark.
Our iSTAR model is built upon the OpenVLA-OFT backbone and follows the same evaluation protocol as
prior work: each task is evaluated with 10 trials, repeated 5 runs.

Overall, iSTAR achieves performance comparable to state-of-the-art VLA models across all four task
suites and consistently outperforms its OpenVLA-OFT backbone.
The improvement margins are moderate, which we attribute to the characteristics of LIBERO:
task instructions are explicit and execution order is largely fixed, leaving limited ambiguity for
additional task-level semantic reasoning.
Under such conditions, strong end-to-end VLA backbones already perform near saturation.
Nevertheless, iSTAR remains competitive on all task suites, including long-horizon tasks, indicating
that in-parameter structural reasoning does not hinder end-to-end execution and can be seamlessly
integrated with strong VLA backbones even when the benefit of explicit reasoning is less pronounced.



\begin{table}[t]
\centering
\caption{Success rate (SR, \%) on four task suites.
Best results are in \textbf{bold}, second-best are \underline{underlined}.}
\label{tab:sr_libero}
\renewcommand{\arraystretch}{1.15}
\setlength{\tabcolsep}{6pt}
\small

\begin{tabular*}{\columnwidth}{@{\extracolsep{\fill}}lccccc}
\toprule
\textbf{Method} & \textbf{Spatial} & \textbf{Object} & \textbf{Goal} & \textbf{Long}  \\
\midrule
OpenVLA \yrcite{DBLP:conf/corl/KimPKXB0RFSVKBT24}  & 84.7 & 88.4 & 79.2 & 53.7   \\
$\pi$0  \yrcite{black2410pi0}              & \textbf{96.8} &\textbf{98.8}  & 95.8 & 85.2   \\
Evo-1  \yrcite{lin2025evo}              &92.7  &97.7  &\textbf{ 96.3} &  \textbf{92.3}  \\
OpenVLA-OFT \yrcite{DBLP:journals/corr/abs-2502-19645}     &\underline{96.2}  &98.3  &\underline{96.2}  &90.7    \\
iSTAR (Ours)     &         \textbf{ 96.8 }  & \underline{ 98.4} & 96.0 &  \underline{ 92.2}  \\
\bottomrule
\end{tabular*}
\vspace{-2em}
\end{table}

\subsection{Real-World Experiments}

To evaluate whether iSTAR generalizes beyond simulation and benchmark settings, we conduct real-world manipulation experiments on an industrial robotic platform. 
These experiments are designed to assess task-level reasoning under realistic control constraints, while minimizing the impact of low-level manipulation difficulty.
Following the protocol of OpenVLA-OFT, we fine-tune the model on our UR3 platform for approximately 50-150K steps with an action chunk size $K=25$. 
At inference time, we adopt a synchronous execution strategy where the full 25-step chunk is executed before the next model re-query. To match the UR3’s $125\text{Hz}$ RTDE control loop, each predicted action in the chunk (originally representing a $40\text{ms}$ interval at $25\text{Hz}$) is internally expanded into 5 sub-steps through linear interpolation, ensuring consistent and fluid mechanical motion.

\begin{figure}[t]
  \centering
  \includegraphics[width=1.0\columnwidth]{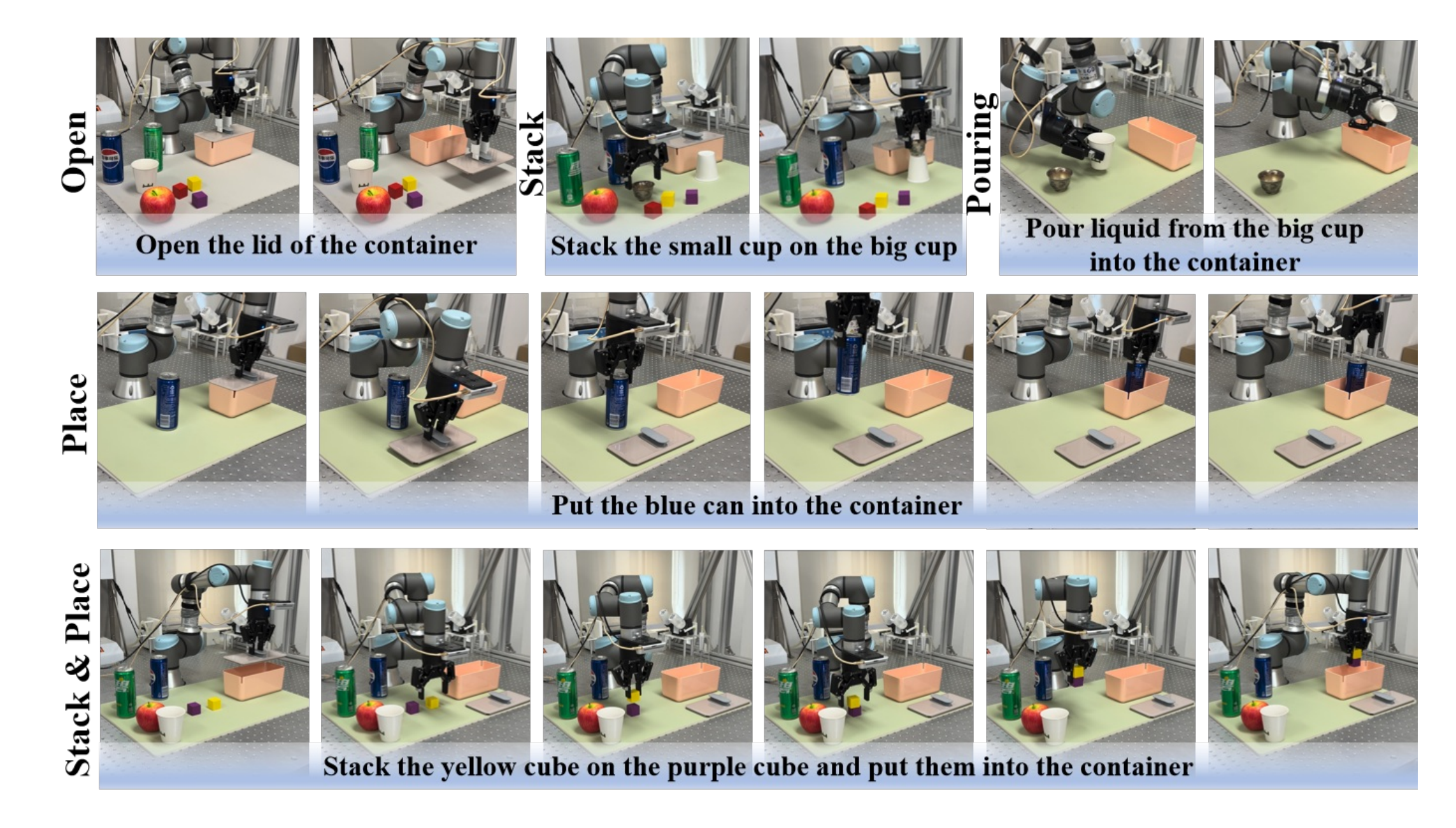}
  \caption{Examples of real-world experiments. For each task, the prompt is shown along with the detailed execution stages, from initial state to final completion.}
  \vspace{-1.5em}
  \label{fig:realworldexample}
\end{figure}

\vspace{-1em}
\paragraph{Task Settings.}
In all tasks, the prompt specifies the desired final state but does not list prerequisite subtasks.
As a result, the same prompt may require different execution structures depending on the initial world state.  Figure \ref{fig:realworldexample} depicts the examples of tasks.

    \noindent \textbf{Open / Close Lid.}  
    Open or close a container lid. \textit{Train}: 20 demos. 
    \textit{Test}: 10 trials.
    

    \noindent \textbf{Stack.}  
    Stack one object onto another, including stacking onto a closed container.
    The prompt does not specify the valid spatial relation. 
    \textit{Train}: 30 demos.
    \textit{Test}: 10 trials.

    \noindent \textbf{Pour.}  
    Pour liquid from one receptacle into another.
    The prompt specifies the transfer goal but not intermediate pouring actions. 
    \textit{Train}: 20 demos. 
    \textit{Test}: 10 trials.

    \noindent \textbf{Place.}  
    Place an object into a container, with the final state optionally requiring the lid to be closed.
    The prompt does not indicate whether opening the lid is necessary. 
    \textit{Train}: 80 demos. 
    \textit{Test}: 30 trials.

    \noindent \textbf{Stack \& Place.}  
    Stack one object onto another and place the stacked objects into a container.
    The prompt does not specify whether the container must be opened before placement. 
    \textit{Train}: 60 demos. 
    \textit{Test}: 20 trials.






\begin{table}[t]
    \centering
    \caption{Real-world task success rates (\%) on the UR3 platform.}
    \label{tab:results}
    \setlength{\tabcolsep}{7pt}
    \small
    \begin{tabular}{lccc}
        \toprule
        \textbf{Task} & \textbf{ OFT+Goal} & \textbf{OFT+Subs} & \textbf{Ours} \\
        \midrule
        Open / Close Lid & 80.0 & 80.0 & \textbf{90.0} \\
        Stack            & 80.0 & 80.0 & \textbf{90.0} \\
        Pouring          & {70.0} & {70.0} & \textbf{80.0} \\
                        Place     & 60.0 & 80.0 & \textbf{86.7} \\
        Stack \& Place   & 45.0 & 55.0 & \textbf{75.0} \\
        \midrule
        \textbf{Average SR} & 67.0 & 73.0 & \textbf{84.3} \\
        \bottomrule
    \end{tabular}
    \vspace{-1em}
\end{table}

\paragraph{Task Performance.}
We report the average success rate (SR) in Table~\ref{tab:results}.
Each task is evaluated with randomized object poses and two different tabletop backgrounds.
We compare our method against the OpenVLA-OFT backbone under two input settings: a goal-only input that specifies only the final task objective (OFT+Goal), and an instruction-following variant that concatenates explicit subtask prompts as input (OFT+Subs), allowing us to isolate the effect of task-level reasoning and generalization.

As shown in Table~\ref{tab:results}, both OFT+Subs and our method improve over the goal-only baseline, indicating that providing subtask-level guidance partially alleviates failures caused by missing prerequisite information.
However, OFT+Subs still struggles to generalize well under variations in initial states and object configurations.
We observe behaviors of redundant actions (e.g., attempting to open an already open lid) and failures when the prescribed subtask sequence does not align with the current scene.
In contrast, by resolving semantic choices in a structured low-dimensional space before action decoding, our approach can reorganize its execution structure when encountering shifted conditions.

\vspace{-1em}
\paragraph{Inference Efficiency.}
Despite incorporating enriched task-level instructions, our model maintains efficient real-time inference.
On the UR3 platform, our implementation achieves an average latency of 0.426s and a throughput of 64.3Hz, supporting reactive closed-loop control.

\section{Conclusion}
We introduced iSTAR, a framework that enhances VLA models by embedding task-level semantic structure directly into model parameters. Through in-parameter structural reasoning and functional differentiation, iSTAR explicitly resolves and reuses task structure before action decoding, without relying on external planners or prompt-based decomposition. 
We provided both theoretical and empirical evidence that such structured semantic commitments improve generalization in long-horizon and ambiguous task settings. Experiments on VIMA-Bench, LIBERO, and real-world robotic manipulation demonstrate more reliable execution and state-conditioned task reorganization, particularly when execution order and prerequisites must be inferred. Overall, iSTAR highlights in-parameter semantic commitment as a principled pathway toward robust and generalizable long-horizon robotic behavior.





\nocite{langley00}

\bibliography{example_paper}
\bibliographystyle{icml2026}

\newpage
\appendix
\onecolumn

\section{Related Work -- Semantic Reasoning from Vision and Language }


In VLA settings, semantic reasoning over compositional and relational structure is critical for high-level decisions.
Such semantic reasoning has been widely studied and shown to be useful in vision–language research.
Early studies on systematic generalization show that learned models tend to exploit statistical regularities in the training distribution, rather than acquiring compositional rules, even when such rules are supported by the training data \cite{DBLP:conf/icml/LakeB18,DBLP:conf/iclr/KeysersSSBFKMSS20}.
Subsequent work suggests that systematic generalization is difficult to achieve without explicit structural priors or inductive bias \cite{DBLP:conf/iclr/BahdanauMNNVC19}.
In the vision–language domain, vision–language researches \cite{DBLP:conf/nips/Yi0G0KT18,hudson2019gqa} explicitly disentangle perception from reasoning, achieving strong robustness and interpretability on VQA benchmarks.
Recent evidence further suggests that even large VLMs struggle to maintain consistent compositional generalization across multiple semantic levels, highlighting the limitations of implicit reasoning in latent representations \cite{li2025consistency}.
Beyond static question answering, structured semantic reasoning has also been explored in spatiotemporal settings, where explicit modeling of semantic relations over time improves action understanding and prediction \cite{wu2021spatial,zhou2024robodreamer}.
Differently, our method treats semantic structure as an explicit commitment for planning, rather than grounding it in generative or visual representations.

\section{Detailed Proofs for Proposition \ref{prop:structured_advantage}}
\label{sec:appendix_proof}

\subsection{Preliminaries and Notation}

We consider a finite-horizon decision problem with horizon \(T\).
Let \(\pi^*\) denote the expert policy and \(\hat\pi\) a learned policy.

At each time step \(t\), the expert admits a latent semantic commitment
\[
c_t^* \in \mathcal C,
\]
\[
p_t^* = r^*(c_t^*),
\]
such that the expert action is sampled as
\[
a_t^* \sim h^*(\cdot \mid I_t, p_t^*).
\]

The structured policy \(\hat\pi\) consists of three components:
(i) a semantic graph predictor \(\hat g : I_{1:T} \mapsto \hat c_{1:T}\),
(ii) a realization map \(\hat r : \mathcal C \to \mathcal P\),
and (iii) an action decoder \(\hat h : (I_t, p_t) \mapsto a_t\).

We measure performance using expected cumulative cost
\[
J(\pi) = \mathbb{E}_{\pi}\!\left[\sum_{t=1}^T \ell(s_t,a_t)\right],
\]
where the per-step loss is bounded as \(\ell \in [0,1]\).

\subsection{Error Decomposition}

We restate the decomposition used in Eq.~(6).

\begin{lemma}[Structured Error Decomposition]
\label{lem:decomposition}
Assume the expert policy admits latent semantic commitments \((c_t^*,p_t^*)\).
Then the performance gap of the structured policy satisfies
\[
J(\pi^*) - J(\hat\pi)
\;\le\;
C_a \,\epsilon_a
\;+\;
C_r \,\epsilon_r
\;+\;
C_G \,\epsilon_G,
\]
where \(\epsilon_a\) is the one-step action decoding error conditioned on the correct realization,
\(\epsilon_r\) is the realization error from semantic graph to subprompt,
and \(\epsilon_G\) is the semantic graph prediction error.
The constants \(C_a,C_r,C_G\) are independent of the horizon \(T\).
\end{lemma}

\paragraph{Proof.}
The result follows the standard performance sensitivity arguments.
We introduce intermediate policies that successively replace learned components with their expert counterparts.
Each replacement induces a bounded deviation in expected cost proportional to the corresponding prediction error.
Summing these deviations yields the stated inequality.
This decomposition does not rely on trajectory-level coupling or no-regret assumptions.
\hfill\(\square\)

\subsection{Horizon-Insensitive Error Terms}

We now formalize why \(\epsilon_a\) and \(\epsilon_r\) do not scale with the horizon.

\begin{lemma}[Action Decoding Error]
\label{lem:action_error}
Conditioned on the correct semantic realization \(p_t^*\), the action decoding error
\[
\epsilon_a
=
\mathbb{E}_{t}\!\left[
\Pr\big(
\hat h(I_t,p_t^*) \neq a_t^*
\big)
\right]
\]
does not accumulate with \(T\).
\end{lemma}

\paragraph{Proof.}
Given \(p_t^*\), the action decoder models a single-step conditional distribution
\((I_t,p_t^*) \mapsto a_t^*\).
This is a standard supervised learning problem whose generalization behavior depends only on the model class and the sample size, not on the rollout horizon.
\hfill\(\square\)

\begin{lemma}[Realization Error]
\label{lem:realization_error}
If the realization map \(\hat r\) is implemented via templates or a constrained grammar, then the realization error \(\epsilon_r\) is bounded by a constant independent of \(T\).
\end{lemma}

\paragraph{Proof.}
Each semantic commitment is realized independently into a subprompt.
Under deterministic or bounded-stochastic realization, realization errors do not propagate across time steps.
\hfill\(\square\)

\subsection{Semantic Graph Prediction Error}

We now analyze the only horizon-sensitive term.

Let \(\mathcal C\) denote the semantic commitment space.
The semantic predictor outputs a sequence \(\hat c_{1:T} \in \mathcal C^T\).

\begin{lemma}[Generalization of Semantic Graph Prediction]
\label{lem:graph_generalization}
Let \(n_G\) be the number of supervised semantic graph samples.
Then with high probability,
\[
\epsilon_G
\;\le\;
\widehat{\epsilon}_G
+
O\!\left(
\sqrt{\frac{T \log|\mathcal C|}{n_G}}
\right),
\]
where \(\widehat{\epsilon}_G\) denotes the empirical semantic prediction error.
\end{lemma}

\paragraph{Proof.}
Predicting semantic commitments over a horizon \(T\) corresponds to a multi-class prediction problem with output space \(\mathcal C^T\).
Applying a union bound over time steps together with standard uniform convergence arguments for discrete hypothesis classes yields the stated bound.
This bound is intentionally loose and serves only to characterize the horizon dependence.
\hfill\(\square\)

Substituting Lemma~\ref{lem:graph_generalization} into Lemma~\ref{lem:decomposition} and absorbing horizon-independent terms yields
\[
J(\pi^*) - J(\hat\pi)
=
O\!\left(
\sqrt{T \log|\mathcal C|}
\right).
\]

\subsection{Comparison with End-to-End Policies}

For an end-to-end policy that directly predicts a length-\(T\) sequence of discrete action tokens, the effective output space is \(\mathcal A^T\).
Standard complexity-based generalization arguments imply
\[
J(\pi^*) - J(\pi_{\mathrm{e2e}})
=
O\!\left(
\sqrt{T \log|\mathcal A|}
\right).
\]

Since \(\log|\mathcal C| < \log|\mathcal A|\), the structured policy admits a strictly tighter leading-order bound for sufficiently large \(T\).
This completes the proof of Proposition~\ref{prop:structured_advantage}.
\hfill\(\square\)

\subsection{Effective Complexity of the Semantic Commitment Space}

The semantic commitment space is constructed from a set of discrete semantic attributes:
\[
\mathcal C = \{(c_1,\dots,c_K)\mid c_k\in\mathcal C_k\}.
\]

Although a semantic commitment can be represented as a tuple, its effective complexity is additive across attributes.
Specifically,
\[
|\mathcal C|
\;=\;
\!
\sum_{k=1}^K  |\mathcal C_k|,
\]
reflecting that each attribute contributes independently to the semantic choice.

This contrasts with the multiplicative complexity \(\prod_k |\mathcal C_k|\), which corresponds to a combinatorial abstraction of executable behaviors and already underestimates the true complexity of the low-level action space \(\mathcal A\).
In practice, \(\mathcal A\) represents discretized physical control over a high-dimensional continuous space, whose effective complexity grows with both control dimensionality and resolution.
Consequently,
\[
\log|\mathcal C| \ll \log|\mathcal A|,
\]
justifying the regime assumed in Proposition~\ref{prop:structured_advantage}.

\section{Model Integration and Architectural Details}

\subsection{Integration with OpenVLA-OFT}

To integrate our method with OpenVLA-OFT, we freeze the first 28 layers of the vision-language model and use them solely as a prompt token embedding extractor.
This design mirrors the role of token embeddings in modular architectures such as VIMA, allowing high-level semantic representations to be reused without altering the backbone's pretrained reasoning capacity.
The remaining layers are reused for action generation, ensuring that low-level execution remains consistent with the original OpenVLA-OFT model.

\subsection{Dynamic Positional Encoding on LIBERO}

OpenVLA-OFT adopts step-level training, where each action chunk is processed independently and temporal structure is not explicitly modeled.
However, when tasks are decomposed into ordered subtasks, awareness of subtask progression is necessary to avoid redundant or inconsistent execution.

To encode subtask progression while remaining compatible with i.i.d.\ chunked training, we construct a lightweight history-aware positional embedding from teacher-forced subtask representations.
For each demonstration, we obtain an ordered subtask sequence $\{p_1,\dots,p_S\}$ and cache the corresponding embeddings $\{e_1,\dots,e_S\}$ extracted from an intermediate language-model layer.
Given a training sample associated with the current subtask index $i$, we summarize the preceding subtasks by an aggregation operator
\begin{equation}
\bar e_{<i} = \mathrm{Agg}\big(\{e_1,\dots,e_{i-1}\}\big),
\end{equation}
and fuse it with the current embedding via a gated update to produce a dynamic positional encoding
\begin{equation}
z_i = \sigma\!\big(W_g [e_i;\bar e_{<i}]\big), \qquad
p_i = z_i \odot e_i + (1-z_i)\odot \bar e_{<i},
\end{equation}
where $\mathrm{Agg}(\cdot)$ can be a simple pooling (e.g., mean/max) or a lightweight prefix encoder.
The resulting $p_i$ conditions action generation (e.g., FiLM-style modulation), providing semantic progress cues without requiring trajectory-level recurrence across chunks.

\section{Subtask Prompt Distillation and Supervision Details}
\label{app:libero_prompt}

This appendix details the construction of subtask prompt supervision for LIBERO, including
subtask prompt generation, demonstration segmentation, visual grounding, and uncertainty-aware
label refinement.

\subsection{Subtask Prompt Set Generation}
\label{app:libero_prompt_set}

LIBERO provides high-level textual task descriptions but does not define explicit subtask-level
semantic structure.
To obtain a canonical set of subtask prompts for each task, we first perform subtask inference
from task descriptions using a pretrained large language model (GPT-4o).

For each LIBERO task, the original textual instruction is provided to the language model, which
is prompted to decompose the task into a small set of semantically meaningful subtasks that
correspond to executable stages (e.g., opening a container, grasping an object, placing it at a
target location).
The resulting subtask prompts form a task-specific subtask vocabulary that is shared across all
demonstrations of the task.

\begin{lstlisting}[caption={Subtask inference prompt used for LIBERO annotation.},
                   label={lst:libero_subtask_prompt}]
You are given a robotic manipulation task from the LIBERO benchmark.
Each task contains a natural language instruction and a formal goal condition.

Your job is to decompose the task into a minimal sequence of atomic subtasks
that correspond to semantically meaningful execution stages.

Guidelines:
- Subtasks should be short, executable action phrases.
- Each subtask should correspond to a single semantic intent.
- Do NOT include low-level motions or control details.
- Do NOT include redundant or implicit steps.
- Only include actions that are necessary to satisfy the given goal.
- Do NOT explain your reasoning.

Below are examples.

=== Example 1 ===
Language: put the frying pan under the cabinet shelf
Goal: (:goal (And (In chefmate_8_frypan_1
                   wooden_two_layer_shelf_1_bottom_region)))

Subtask:
1. pick up the frying pan and place it under the cabinet shelf

=== Example 2 ===
Language: put the white bowl on top of the cabinet
Goal: (:goal (And (On white_bowl_1
                   wooden_two_layer_shelf_1_top_side)))

Subtask:
1. pick up the white bowl and place it on top of the cabinet

=== Example 3 ===
Language: turn on the stove and put the frying pan on it
Goal: (:goal (And (Turnon flat_stove_1)
                  (On chefmate_8_frypan_1
                      flat_stove_1_cook_region)))

Subtask:
1. turn on the stove
2. pick up the frying pan and place it on the stove

=== Task ===
Language: {TASK_LANGUAGE}
Goal: {TASK_GOAL}

Subtask:
\end{lstlisting}

\subsection{Demonstration Segmentation and Keyframe Extraction}
\label{app:libero_segmentation}

Each LIBERO demonstration is segmented into subtask-level intervals based on gripper state
changes and action transitions.
Specifically, we detect grasp and release events from the gripper state and use them as anchors
to partition a trajectory into semantically coherent segments.

\begin{figure}[t]
  \centering
  \includegraphics[width=\linewidth]{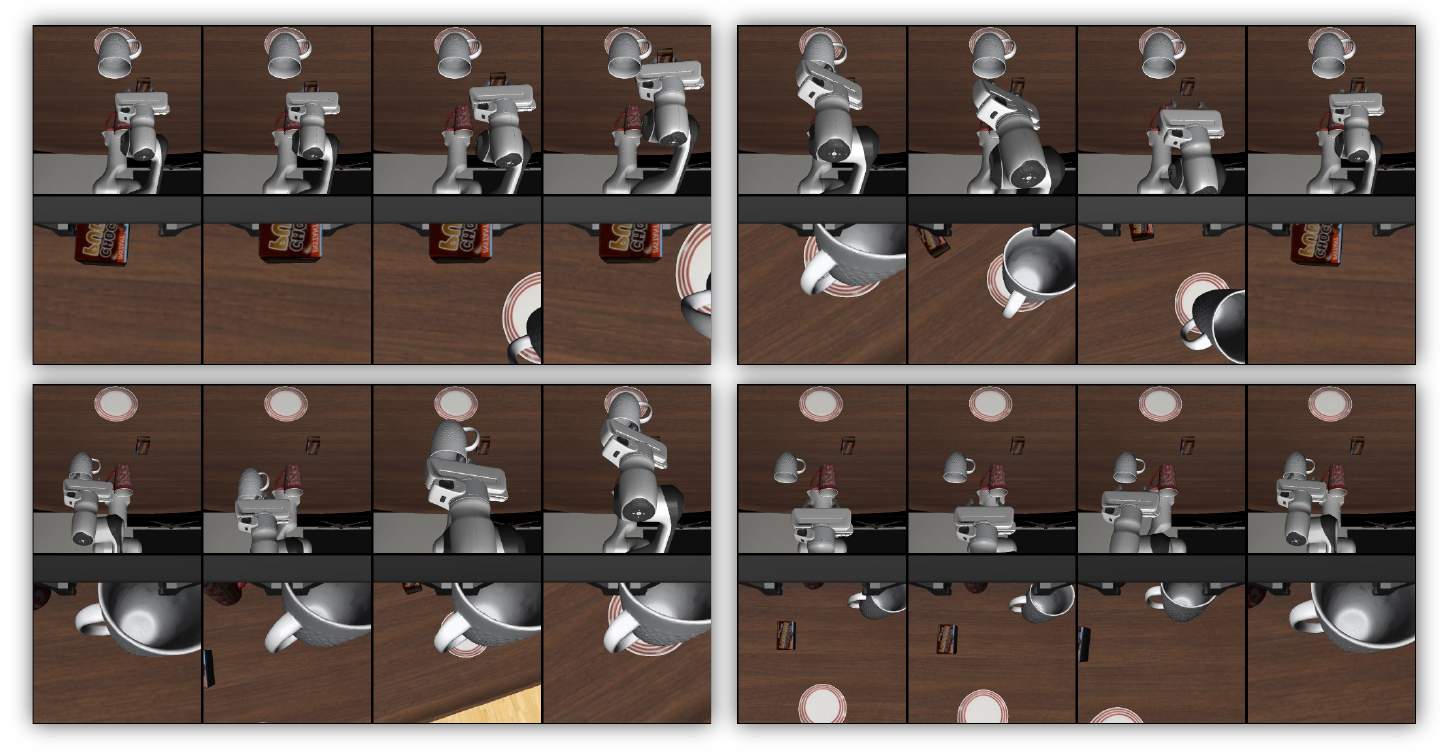}
  \caption{
  Example of demonstration segmentation and visual subtask classification.
  A demonstration of the task
  ``put the white mug on the plate and put the chocolate pudding to the right of the plate''
  is segmented into four subtask-level segments.
  For each segment, eight representative keyframes are extracted and independently
  classified into one of the candidate subtasks.
  Segment-level predictions are merged in temporal order to form the final subtask sequence.
  }
  \label{fig:libero_segmentation_example}
\end{figure}

For every segment, we extract a fixed number of keyframes (eight in our implementation, see Figure \ref{fig:libero_segmentation_example})
that summarize the visual context of the segment.
Subtask classification is performed independently for each segment by conditioning a
vision-language model on the keyframes and a predefined set of candidate subtasks.

For robustness, each segment is classified multiple times with independent sampling.
The final subtask label for a segment is determined by majority voting.
Segment-level predictions are then merged in temporal order to obtain a subtask sequence
for the full demonstration.

\subsection{Visual Subtask Classification via GPT-4o}
\label{app:libero_classification}

Given a segmented demonstration and the task-specific subtask prompt set, we classify each
segment by querying GPT-4o with the extracted keyframes and the candidate subtask list.
The model is instructed to assign the most likely subtask label based solely on the visual
evidence and the predefined subtask vocabulary.

To estimate prediction uncertainty, each classification query is repeated five times with
independent sampling.
We compute an uncertainty score based on label disagreement across samples.
Segments with low uncertainty are automatically assigned the predicted subtask label.

\paragraph{Classification prompt.}
The prompt used for visual subtask classification is shown below.

\begin{lstlisting}[caption={Visual subtask classification prompt used for segment-level annotation.},
                   label={lst:visual_subtask_prompt}]
You are a robotic perception and task grounding model.

Given:
1. The current RGB observation from the robot.
2. A list of candidate subtasks.

Image structure:
- The input image is a composite image arranged in a grid.
- The TOP ROW shows images from the global (scene-level) camera.
- The BOTTOM ROW shows images from the end-effector (egocentric) camera.
- Each COLUMN corresponds to the same time step.
- Images in the same column are synchronized.

View usage:
- Use the GLOBAL view to judge object states and relations (e.g., on/off, on top of).
- Use the END-EFFECTOR view to judge active manipulation or physical contact.

Your task:
Select the ONE subtask that best matches the CURRENT visual state.

Critical rules:
- Choose ONLY ONE subtask from the list.
- Do NOT infer task order or future steps.
- Do NOT select a subtask whose required object state is NOT visible.
- If a subtask is not currently happening or its preconditions are not met, it must be rejected.
- Base the decision strictly on visible object states and relations.
- The RIGHTMOST column corresponds to the FINAL time step.
- The decision should focus primarily on the images in the FINAL column.

Subtasks:
{text}

Output format:
Selected subtask: <copy the exact subtask text>
\end{lstlisting}

\subsection{Uncertainty-Aware Manual Refinement}
\label{app:libero_refinement}

Segments with high classification uncertainty are flagged for manual inspection.
In practice, these cases typically correspond to visually ambiguous transitions or rare execution
patterns.
A lightweight manual correction step is applied to these segments to ensure label consistency
and semantic correctness.

This uncertainty-aware refinement strategy significantly reduces annotation effort while
maintaining high-quality subtask supervision.
In our experiments, only a small fraction of segments require manual correction.

\subsection{Summary}
\label{app:libero_summary}

Through task-level subtask inference, execution-aligned segmentation, visual classification, and
uncertainty-aware refinement, we construct reliable subtask prompt supervision for LIBERO without
requiring dense manual annotation.
This supervision enables structured semantic reasoning within iSTAR while remaining compatible
with standard VLA training pipelines.

\begin{table*}[h]
\centering
\caption{Detailed L1-level results across different task categories.}
\small
\label{tab:l1_transpose}
\begin{tabular}{lccccccc}
\toprule
Task
& \makecell{VIMA \\ {\scriptsize(Small)}}
& \makecell{VIMA \\ {\scriptsize(Large)}}
& \makecell{iSTAR \\ {\scriptsize(Stacked VIMA)}}
& \makecell{iSTAR \\ {\scriptsize(w/o Graph)}}
& \makecell{iSTAR \\ {\scriptsize(w/o Order Entropy)}}
& \makecell{iSTAR \\ {\scriptsize(w/o Concept Alignment)}}
& iSTAR \\
\midrule
follow\_order & 0.75 & 0.84 & 0.77 & 0.82 & 0.82 & 0.81 & 0.87 \\
manipulate\_old\_neighbor & 0.43 & 0.47 & 0.44 & 0.48 & 0.50 & 0.50 & 0.50 \\
novel\_adj & 0.99 & 0.99 & 0.99 & 1.00 & 1.00 & 1.00 & 1.00 \\
novel\_noun & 0.99 & 0.99 & 0.99 & 1.00 & 1.00 & 1.00 & 1.00 \\
pick\_in\_order\_then\_restore & 0.61 & 0.57 & 0.60 & 0.63 & 0.66 & 0.70 & 0.68 \\
rearrange & 0.73 & 0.65 & 0.75 & 0.72 & 0.75 & 0.79 & 0.81 \\
rearrange\_then\_restore & 0.16 & 0.14 & 0.14 & 0.18 & 0.18 & 0.25 & 0.22 \\
rotate & 1.00 & 1.00 & 0.99 & 0.99 & 1.00 & 0.98 & 1.00 \\
same\_shape & 0.95 & 0.98 & 0.95 & 0.96 & 0.98 & 0.97 & 0.99 \\
scene\_understanding & 1.00 & 1.00 & 1.00 & 1.00 & 1.00 & 1.00 & 1.00 \\
twist & 0.09 & 0.24 & 0.09 & 0.18 & 0.25 & 0.22 & 0.23 \\
visual\_manipulation & 1.00 & 1.00 & 1.00 & 1.00 & 1.00 & 1.00 & 1.00 \\
\midrule
Average & 0.724 & 0.739 & 0.724 & 0.748 & 0.764 & 0.778 & 0.786 \\
\bottomrule
\end{tabular}
\end{table*}

\begin{table*}[h]
\centering
\caption{Detailed L2-level results across different task categories.}
\small
\label{tab:l2_transpose}
\begin{tabular}{lccccccc}
\toprule
Task
& \makecell{VIMA \\ {\scriptsize(Small)}}
& \makecell{VIMA \\ {\scriptsize(Large)}}
& \makecell{iSTAR \\ {\scriptsize(Stacked VIMA)}}
& \makecell{iSTAR \\ {\scriptsize(w/o Graph)}}
& \makecell{iSTAR \\ {\scriptsize(w/o Order Entropy)}}
& \makecell{iSTAR \\ {\scriptsize(w/o Concept Alignment)}}
& iSTAR \\
\midrule

follow\_order & 0.81 & 0.78 & 0.81 & 0.88 & 0.90 & 0.89 & 0.94 \\
manipulate\_old\_neighbor & 0.48 & 0.47 & 0.45 & 0.47 & 0.51 & 0.49 & 0.50 \\
novel\_adj & 0.99 & 1.00 & 1.00 & 1.00 & 1.00 & 1.00 & 1.00 \\
novel\_noun & 1.00 & 1.00 & 1.00 & 1.00 & 1.00 & 1.00 & 1.00 \\
pick\_in\_order\_then\_restore & 0.61 & 0.63 & 0.64 & 0.70 & 0.70 & 0.72 & 0.73 \\
rearrange & 0.74 & 0.76 & 0.74 & 0.75 & 0.80 & 0.82 & 0.85 \\
rearrange\_then\_restore & 0.16 & 0.24 & 0.13 & 0.19 & 0.19 & 0.21 & 0.24 \\
rotate & 1.00 & 0.99 & 1.00 & 1.00 & 1.00 & 1.00 & 1.00 \\
same\_shape & 0.98 & 0.96 & 1.00 & 0.99 & 1.00 & 1.00 & 1.00 \\
scene\_understanding & 1.00 & 1.00 & 0.97 & 0.99 & 0.99 & 1.00 & 1.00 \\
twist & 0.09 & 0.11 & 0.09 & 0.15 & 0.22 & 0.17 & 0.24 \\
visual\_manipulation & 1.00 & 1.00 & 1.00 & 1.00 & 1.00 & 1.00 & 1.00 \\
\midrule
Average & 0.738 & 0.745 & 0.736 & 0.760 & 0.776 & 0.775 & 0.792 \\
\bottomrule
\end{tabular}
\end{table*}


\begin{table*}[h]
\centering
\caption{Detailed L3-level results across different task categories.}
\small
\label{tab:l3_transpose}
\begin{tabular}{lccccccc}
\toprule
Task
& \makecell{VIMA \\ {\scriptsize(Small)}}
& \makecell{VIMA \\ {\scriptsize(Large)}}
& \makecell{iSTAR \\ {\scriptsize(Stacked VIMA)}}
& \makecell{iSTAR \\ {\scriptsize(w/o Graph)}}
& \makecell{iSTAR \\ {\scriptsize(w/o Order Entropy)}}
& \makecell{iSTAR \\ {\scriptsize(w/o Concept Alignment)}}
& iSTAR \\
\midrule
follow\_order & 0.80 & 0.88  & 0.89 & 0.92 & 0.90 & 0.94 & 0.96 \\
manipulate\_old\_neighbor & 0.46 & 0.38  & 0.40 & 0.44 & 0.55 & 0.53 & 0.55 \\
novel\_adj & 0.98 & 0.99  & 1.00 & 1.00 & 1.00 & 1.00 & 1.00 \\
novel\_noun & 0.99 & 0.99  & 1.00 & 1.00 & 1.00 & 1.00 & 1.00 \\
pick\_in\_order\_then\_restore & 0.36 & 0.32  & 0.40 & 0.51 & 0.51 & 0.54 & 0.55 \\
rearrange & 0.69 & 0.70  & 0.66 & 0.75 & 0.71 & 0.76 & 0.79 \\
rearrange\_then\_restore & 0.16 & 0.22  & 0.23 & 0.25 & 0.25 & 0.29 & 0.28 \\
rotate & 1.00 & 0.99  & 1.00 & 1.00 & 1.00 & 1.00 & 1.00 \\
same\_shape & 0.95 & 0.95  & 0.94 & 0.95 & 0.99 & 0.95 & 1.00 \\
scene\_understanding & 0.98 & 0.99  & 1.00 & 1.00 & 1.00 & 1.00 & 1.00 \\
twist & 0.17 & 0.16  & 0.17 & 0.17 & 0.25 & 0.23 & 0.27 \\
visual\_manipulation & 0.99 & 0.99  & 1.00 & 1.00 & 1.00 & 1.00 & 1.00 \\
\midrule
Average & 0.711 & 0.720 & 0.724 & 0.749 & 0.763 & 0.770 & 0.783 \\
\bottomrule
\end{tabular}
\end{table*}

\begin{table*}[h]
\centering
\caption{Detailed L4-level results across different task categories.}
\small
\label{tab:l4_transpose}
\begin{tabular}{lccccccc}
\toprule
Task
& \makecell{VIMA \\ {\scriptsize(Small)}}
& \makecell{VIMA \\ {\scriptsize(Large)}}
& \makecell{iSTAR \\ {\scriptsize(Stacked VIMA)}}
& \makecell{iSTAR \\ {\scriptsize(w/o Graph)}}
& \makecell{iSTAR \\ {\scriptsize(w/o Order Entropy)}}
& \makecell{iSTAR \\ {\scriptsize(w/o Concept Alignment)}}
& iSTAR \\
\midrule

follow\_motion & 0.00 & 0.00 & 0.00 & 0.01 & 0.05 & 0.05 &0.06\\
novel\_adj\_and\_noun & 0.10 & 0.85 & 0.09 & 0.15 & 0.23 & 0.18 & 0.24\\
same\_texture & 0.95 & 0.92 & 0.93 & 0.97 & 1.00 & 0.97 &1.00\\
\midrule
Average & 0.350 & 0.590 & 0.340 & 0.377 & 0.427 & 0.400 &0.433\\
\bottomrule
\end{tabular}
\end{table*}



\section{Extended Ablation Studies}
\label{app:extend_vima}

We report additional ablation results examining alternative design choices, including static positional encoding and different subtask selection strategies.
These results further confirm the necessity of state-conditioned subtask reasoning and dynamic temporal modeling.

\paragraph{L1--L4: Four levels of zero-shot generalization.}
VIMA-BENCH defines a four-level evaluation protocol in which each level deviates further from the training distribution and is therefore harder than the previous one.

\begin{itemize}
  \item \textbf{L1: Placement generalization.}
  All prompt templates are seen during training; at test time, only the object placement on the tabletop is randomized.

  \item \textbf{L2: Combinatorial generalization.}
  All objects and textures are seen during training, but novel combinations of them appear at test time.

  \item \textbf{L3: Novel object generalization.}
  The test prompts and the simulated workspace include novel objects and textures that were not seen in training.

  \item \textbf{L4: Novel task generalization.}
  Test time introduces new tasks specified by novel prompt templates (i.e., unseen task templates).
\end{itemize}

We report detailed success rates for each task across different VIMA task-level groups, as shown in Tables~\ref{tab:l1_transpose}--\ref{tab:l4_transpose}.
Across long-horizon tasks such as \textit{follow\_order} and \textit{rearrange}, our method consistently achieves substantial improvements over the baselines.
These tasks require reasoning over prerequisite satisfaction and execution order, highlighting the advantage of state-conditioned subtask reasoning.
In contrast, performance on the \textit{twist} task remains significantly lower across all methods.
Although the underlying motor primitive in \textit{twist} is identical to that in \textit{rotate}, the two tasks differ critically in their specification.
While \textit{rotate} explicitly provides a target rotation angle, \textit{twist} only specifies the desired post-rotation visual appearance of the object.
As a result, successful execution of \textit{twist} requires accurate visual inference of object orientation and precise grounding of this inferred state into action commands.
This discrepancy suggests that a major challenge for VLA models lies in the tight coupling between visual reasoning and action execution.
Estimating object rotation angles from visual observations demands strong perceptual and geometric reasoning capabilities.
To further investigate this issue, we conducted preliminary experiments using both traditional computer vision pipelines and learning-based approaches to directly estimate object rotation angles.
The results indicate that these methods achieve accuracies of only around 50\%, suggesting that the difficulty of the \textit{twist} task may exceed the current capacity of the models and perceptual tools employed in this work.

We further investigate the effect of different subprompt projector designs.
Tables~\ref{tab:l1_projector}--\ref{tab:l4_projector} report results using alternative projector architectures across different VIMA task-level groups.
In addition to our default encoder-based design, denoted as \textit{iSTAR (Enc)}, which predicts all subprompt embeddings in a single forward pass, we evaluate an encoder--decoder variant, \textit{iSTAR (Enc--Dec)}.
The encoder--decoder model generates subprompt embeddings autoregressively, introducing an explicit temporal dependency among subprompts.
Empirically, we observe that \textit{iSTAR (Enc--Dec)} yields slightly inferior performance compared to the encoder-only variant across most task groups.
A plausible explanation is that the additional autoregressive structure introduces unnecessary modeling complexity for the considered subprompt space.
In our setting, subprompts are relatively simple and template-based, and predicting them jointly with a shared encoder appears sufficient and more parameter-efficient.

\begin{table*}[h]
\centering
\caption{Comparison of different subprompt projectors on the L1 set.}
\label{tab:l1_projector}
\begin{tabular}{lcccc}
\toprule
Task & VIMA (Small) & VIMA (Large) & iSTAR (Enc-Dec) & iSTAR (Enc) \\
\midrule
follow\_order & 0.75 & 0.84 & 0.77 & 0.82 \\
manipulate\_old\_neighbor & 0.43 & 0.47 & 0.44 & 0.48 \\
novel\_adj & 0.99 & 0.99 & 0.99 & 1.00 \\
novel\_noun & 0.99 & 0.99 & 0.99 & 1.00 \\
pick\_in\_order\_then\_restore & 0.61 & 0.57 & 0.65 & 0.63 \\
rearrange & 0.73 & 0.65 & 0.67 & 0.72 \\
rearrange\_then\_restore & 0.16 & 0.14 & 0.14 & 0.18 \\
rotate & 1.00 & 1.00 & 0.99 & 0.99 \\
same\_shape & 0.95 & 0.98 & 0.97 & 0.96 \\
scene\_understanding & 1.00 & 1.00 & 1.00 & 1.00 \\
twist & 0.09 & 0.24 & 0.11 & 0.18 \\
visual\_manipulation & 1.00 & 1.00 & 1.00 & 1.00 \\
\midrule
Average & 0.724 & 0.739 & 0.723 & 0.748 \\
\bottomrule
\end{tabular}
\end{table*}

\begin{table*}[h]
\centering
\caption{Comparison of different subprompt projectors on the L2 set.}
\label{tab:l2_projector}
\begin{tabular}{lcccc}
\toprule
Task & VIMA (Small) & VIMA (Large) & iSTAR (Enc-Dec) & iSTAR (Enc) \\
\midrule
follow\_order & 0.81 & 0.78 & 0.80 & 0.88 \\
manipulate\_old\_neighbor & 0.48 & 0.47 & 0.45 & 0.47 \\
novel\_adj & 0.99 & 1.00 & 0.99 & 1.00 \\
novel\_noun & 1.00 & 1.00 & 1.00 & 1.00 \\
pick\_in\_order\_then\_restore & 0.61 & 0.63 & 0.65 & 0.70 \\
rearrange & 0.74 & 0.76 & 0.78 & 0.75 \\
rearrange\_then\_restore & 0.16 & 0.24 & 0.22 & 0.19 \\
rotate & 1.00 & 0.99 & 1.00 & 1.00 \\
same\_shape & 0.98 & 0.96 & 0.94 & 0.99 \\
scene\_understanding & 1.00 & 1.00 & 1.00 & 0.99 \\
twist & 0.09 & 0.11 & 0.09 & 0.15 \\
visual\_manipulation & 1.00 & 1.00 & 0.99 & 1.00 \\
\midrule
Average & 0.737 & 0.745 & 0.743 & 0.760 \\
\bottomrule
\end{tabular}
\end{table*}

\begin{table*}[h]
\centering
\caption{Comparison of different subprompt projectors on the L3 set.}
\label{tab:l3_projector}
\begin{tabular}{lcccc}
\toprule
Task & VIMA (Small) & VIMA (Large) & iSTAR (Enc-Dec) & iSTAR (Enc) \\
\midrule
follow\_order & 0.80 & 0.88 & 0.80 & 0.92 \\
manipulate\_old\_neighbor & 0.46 & 0.38 & 0.42 & 0.44 \\
novel\_adj & 0.98 & 0.99 & 1.00 & 1.00 \\
novel\_noun & 0.99 & 0.99 & 1.00 & 1.00 \\
pick\_in\_order\_then\_restore & 0.36 & 0.32 & 0.44 & 0.51 \\
rearrange & 0.69 & 0.70 & 0.72 & 0.75 \\
rearrange\_then\_restore & 0.16 & 0.22 & 0.21 & 0.25 \\
rotate & 1.00 & 0.99 & 1.00 & 1.00 \\
same\_shape & 0.95 & 0.95 & 0.95 & 0.95 \\
scene\_understanding & 0.98 & 0.99 & 1.00 & 1.00 \\
twist & 0.17 & 0.16 & 0.18 & 0.17 \\
visual\_manipulation & 0.99 & 0.99 & 1.00 & 1.00 \\
\midrule
Average & 0.716 & 0.720 & 0.726 & 0.751 \\
\bottomrule
\end{tabular}
\end{table*}

\begin{table*}[h]
\centering
\caption{Comparison of different subprompt projectors on the L4 set.}
\label{tab:l4_projector}
\begin{tabular}{lcccc}
\toprule
Task & VIMA (Small) & VIMA (Large) & iSTAR (Enc-Dec) & iSTAR (Enc) \\
\midrule
follow\_motion & 0.00 & 0.00 & 0.00 & 0.01 \\
novel\_adj\_and\_noun & 0.10 & 0.85 & 0.11 & 0.15 \\
same\_texture & 0.95 & 0.92 & 0.92 & 0.97 \\
\midrule
Average & 0.350 & 0.590 & 0.343 & 0.377 \\
\bottomrule
\end{tabular}
\end{table*}

\begin{figure}[h]
  \centering
  \includegraphics[width=0.9\columnwidth]{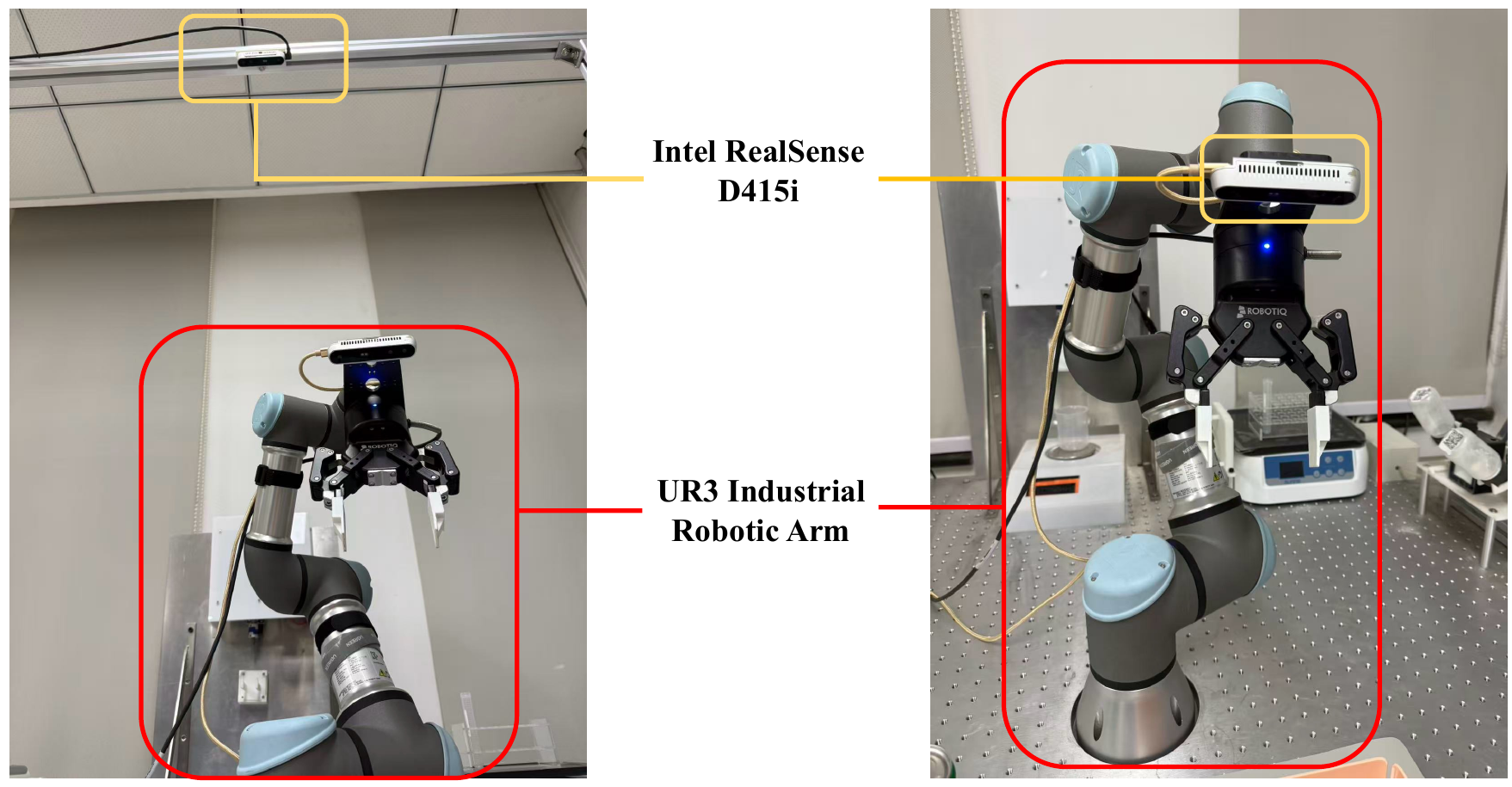}
  \caption{Hardware platform and sensing in our real-world experiments: one UR3 industrial robotic arm and two Intel RealSense D415i cameras. The left figure shows an overview of the experimental hardware platform, and the right figure provides a close-up view of the robotic arm.}
  \label{fig:realworldexample2}
\end{figure}

\section{Additional Real-World Experimental Details}

\subsection{Hardware Platform and Sensing}

All real-world experiments are conducted on a UR3 industrial robotic arm with 6 degrees of freedom, equipped with a parallel-jaw gripper.
The system provides two visual observation streams: a wrist-mounted eye-in-hand camera capturing first-person views during manipulation, and an overhead third-person camera (Intel RealSense D415i) providing a global scene view.
In addition to visual inputs, the policy receives proprioceptive robot state observations consisting of the 7-dimensional joint angles and gripper width.
This multi-view sensing configuration enables both local object interaction and global scene understanding during task execution.

\subsection{Control Interface and Action Execution}

The UR3 robot operates at a native control frequency of 125 Hz via the RTDE interface.
To maintain consistency with the OpenVLA-OFT backbone, our VLA policy predicts actions at 25 Hz using action chunking.
At inference time, a synchronous execution strategy is adopted, where each predicted action chunk is fully executed before querying the model again.

To bridge the frequency mismatch between the policy and the robot controller, we employ interpolation to upsample the predicted actions.
Specifically, each action in the 25 Hz chunk, corresponding to a 40 ms interval, is expanded into five evenly spaced sub-steps, resulting in smooth and stable motion at the robot’s native control rate.
This interpolation is applied purely at the execution layer and does not alter the policy architecture or training procedure.

\subsection{Task Setting Details}

For all real-world tasks, the language prompt specifies the desired final state of the task, such as the target object configuration or whether a container lid should be closed at completion.
However, the prompt does not enumerate prerequisite subtasks or intermediate actions.
As a result, the execution structure required to reach the specified goal depends on the initial world state.
This design ensures that task completion requires inferring valid action structure conditioned on the scene, rather than following explicitly provided procedural instructions.
We describe the tasks and summarize the corresponding prompts and subprompts in Table~\ref{tab:prompt_subprompt}.

\begin{table}[h]
\centering
\caption{Summary of real-world tasks with corresponding prompts and subprompts.
Prompts specify the desired final state, while subprompts represent intermediate semantic goals that must be inferred based on the initial world state.}
\label{tab:prompt_subprompt}
\setlength{\tabcolsep}{3pt}
\begin{tabular}{p{2.4cm} p{2.4cm} p{4.1cm} p{4.3cm} c c}
\toprule
\textbf{Task} & \textbf{Initial State} & \textbf{Prompt (Goal)} & \textbf{Subprompt(s)} & \textbf{\# Demos} & \textbf{\# Trials} \\
\midrule



Open / Close
& Lid closed
& Open the lid 
& Open the lid 
& 10
& 5 \\

& Lid closed
& Close the lid
& Close the lid
& 10
& 5 \\

\midrule

Stack
& -
& Stack X on Y
& Stack X on Y
& 30
& 10 \\

\midrule

Pouring
& -
& Pour liquid from cup A into cup B
& Pour liquid from cup A into cup B
& 10
& 5 \\

& -
& Pour liquid from cup into container
& Pour liquid from cup into container
& 10
& 5 \\

\midrule

Place
& Lid closed
& Put X into the container and close the lid
& Open the lid; put X into the container; close the lid
& 20
& 10 \\

& Lid closed
& Put X into the container
& Open lid; put X into container
& 30
& 10 \\

& Lid open
& Put X into the container
& Put X into container
& 30
& 10 \\

\midrule

Stack \& Place
& Lid closed
& Stack the X on Y and put them into the container
& Stack X on Y; open lid; put the stacked objects into the container
& 30
& 10 \\

& Lid open
& Stack X on Y and put them into the container
& Stack X on Y; put the stacked objects into the container
& 30
& 10 \\
\bottomrule
\end{tabular}
\end{table}

\subsection{Task-Specific Success Rate}

We show the task-specific results on the real-world experiment in Table \ref{tab:detailed_sr}.

\subsection{Comparative Analysis of Initial State Robustness}
\label{app:breakdown}

To understand the performance gap between instruction-following baselines and our proposed method, we evaluate Task (Place) and Task (Stack \& Place) under two distinct initial world states: Lid Closed and Lid Open. While OFT+Subs benefits from fine-grained subtask prompts, it lacks the visual closed-loop adaptability required to reorganize its plan when prerequisites are already satisfied.

To provide a granular assessment of policy performance, we implement a staged scoring system. Each task is decomposed into semantic milestones, with points awarded cumulatively. This structure is designed to isolate the performance of the high-level reasoning branch from low-level motor control.

\paragraph{Place Object into Container (100 pts)} \begin{itemize} \item \textbf{Stage 1 (20 pts):} \textit{Reaching} --- End-effector successfully approaches the target object. \item \textbf{Stage 2 (20 pts):} \textit{Grasping} --- Securely lifting the object and maintaining stability. \item \textbf{Stage 3 (30 pts):} \textbf{\textit{Reasoning Branch (Lid Constraint)}} --- \begin{itemize} \item \textbf{Condition A (Lid Closed):} Successfully executes the ``Open Lid" subtask. \item \textbf{Condition B (Lid Open):} Corrects the plan by bypassing the ``Open" skill and moving directly to the container. \end{itemize} \item \textbf{Stage 4 (30 pts):} \textit{Final Placement} --- Releasing the object in the right place. \end{itemize}

\paragraph{Stack and Place into Container (100 pts)} \begin{itemize} \item \textbf{Stage 1 (20 pts):} \textit{Nesting} --- Accurate stacking objects. \item \textbf{Stage 2 (20 pts):} \textit{Assembly Transport} --- Lifting the nested assembly and moving towards the box. \item \textbf{Stage 3 (30 pts):} \textbf{\textit{Reasoning Branch (Lid Constraint)}} --- \begin{itemize} \item \textbf{Condition A (Lid Closed):} Successfully executes the ``Open Lid" subtask. \item \textbf{Condition B (Lid Open):} Corrects the plan by bypassing the ``Open" skill and moving directly to the container. \end{itemize} \item \textbf{Stage 4 (30 pts):} \textit{Final Placement} --- Releasing the object in the right place. \end{itemize}

\begin{table}[h]
    \centering
    \caption{Detailed Success Rates (\%) by Task Variant and Execution Structure on the UR3 platform. We compare our dynamic reasoning approach against \textit{OFT+Goal} and the instruction-following \textit{OFT+Subs} baseline.}
    \label{tab:detailed_sr}
    \setlength{\tabcolsep}{12pt}
    \begin{tabular}{ll ccc}
        \toprule
        \textbf{Task} & \textbf{Variant / Execution Structure} & \textbf{OFT+Goal} & \textbf{OFT+Subs} & \textbf{Ours} \\
        \midrule
        \multirow{2}{*}{Open / Close Lid} & Open Lid & 80.0 & 80.0 & \textbf{100.0} \\
                                            & Close Lid & 80.0 & 80.0 & \textbf{80.0} \\
       
        \midrule
        \multirow{1}{*}{Stack}           & On Object & 80.0 & 80.0 & \textbf{90.0} \\

        \midrule        \multirow{2}{*}{Pour}         & Cup-to-Cup & 60.0 & 60.0 & \textbf{80.0} \\
                                            & Cup-to-Container &80.0 & 80.0 & \textbf{80.0} \\

                                             \midrule
        \multirow{3}{*}{Place}    & Open $-$ Place $-$ Close & 40.0 & 80.0 & \textbf{90.0} \\
                                            & Open $-$ Place & 60.0 & 80.0 & \textbf{80.0} \\
                                            & Place & 80.0 & 80.0 & \textbf{90.0} \\
        \midrule
        \multirow{2}{*}{Stack \& Place}  & Stack $-$ Open $-$ Place & 30.0 & 50.0 & \textbf{70.0} \\
                                            & Stack $-$ Place & 60.0 & 60.0& \textbf{80.0} \\
                                            
        \midrule
        \textbf{Average SR}                 & --- & 67.0 & 73.0 & \textbf{84.3} \\

        \bottomrule
    \end{tabular}
\end{table}

Table~\ref{tab:detailed_breakdown} provides a fine-grained analysis of policy behavior under different
initial world states by decomposing each task into semantic stages.
Across both \textit{Place} and \textit{Stack \& Place} tasks, low-level motor stages (Stage~1,2 and
Stage~4) show relatively small performance gaps among methods, indicating that execution-level
control is not the primary source of failure.

In contrast, the largest and most consistent performance differences appear in
\textbf{Stage~3 (Reasoning Branch)}, which requires adapting the execution plan based on the initial
lid state.
When the lid is already open, both OFT+Goal and OFT+Subs frequently fail to bypass the redundant
``Open Lid'' subtask, leading to unnecessary actions and cascading failures.
Although OFT+Subs benefits from subtask supervision, it still lacks the ability to reorganize its
plan based on visual feedback.

Our method consistently achieves near-optimal scores in Stage~3 under both lid conditions,
demonstrating robust state-conditioned task reorganization.
These results confirm that the performance gains primarily stem from improved task-level reasoning
and closed-loop adaptation, highlighting the advantage of
explicit semantic reasoning for handling varying initial states.

\begin{table}[H]
    \centering
    \caption{Detailed performance breakdown (average scores) across task stages on the UR3 platform. Scores are cumulative (100 pts total per task). We highlight the performance drop in \textbf{Stage 3 (Reasoning Branch)}, where baselines struggle to adapt to varying initial lid states.}
    \label{tab:detailed_breakdown}
    \setlength{\tabcolsep}{6pt}
    \begin{tabular}{ll cccc c}
        \toprule
        \textbf{Task \& Initial State} & \textbf{Method} & \textbf{Stage 1} & \textbf{Stage 2} & \textbf{Stage 3} & \textbf{Stage 4} & \textbf{Total} \\
        & & (Reach/Nest) & (Grasp/Trans) & \textbf{(Reasoning)} & (Placement) & \textbf{(Score)} \\
        \midrule
        \textbf{Place \& Close } & OFT+Goal & 15 & 14 & 15 & 12 & 56 \\
        \textit{(Lid Closed)} & OFT+Subs & 19 & 18 & 27 & 24 & 88 \\
        & \textbf{Ours} & \textbf{20} & \textbf{20} & \textbf{30} & \textbf{27} & \textbf{97} \\
        \midrule
        \textbf{Place } & OFT+Goal & 17 & 17 & 18 & 18 & 70 \\
        \textit{(Lid Closed)} & OFT+Subs & 19 & 18 & 27 & 24 & 88 \\
        & \textbf{Ours} & \textbf{19} & \textbf{19} & \textbf{30} & \textbf{24} & \textbf{92} \\
        \midrule
        \textbf{Place } & OFT+Goal  & 18 & 18 & 24 & 24 & 84 \\
        \textit{(Lid Open)} & OFT+Subs & 18 & 18 & 24 & 24 & 84 \\
        & \textbf{Ours} & \textbf{19} & \textbf{19} & \textbf{27} & \textbf{27} & \textbf{92} \\
        \midrule
        \textbf{Stack \& Place} & OFT+Goal & 16 & 14 & 12 & 9 & 51 \\
        \textit{(Lid Closed)} & OFT+Subs & 18 & 16 & 21 & 15 & 70 \\
        & \textbf{Ours} & \textbf{18} & \textbf{16} & \textbf{27} & \textbf{21} & \textbf{82} \\
        \midrule
        \textbf{Stack \& Place} & OFT+Goal & 16 & 14 & 18 & 18 & 66 \\
    \textit{(Lid Open)} & OFT+Subs & \textbf{18} & 16 & 24 & 18 & 76 \\
        & \textbf{Ours} & {16} & \textbf{20} & \textbf{30} & \textbf{24} & \textbf{90} \\
        \bottomrule
    \end{tabular}
\end{table}

\section{Visualization}

In this appendix, we provide qualitative visualizations to complement the quantitative evaluations in the main paper.
Each visualization presents the full task prompt, the subprompts, and step-wise execution snapshots, enabling a transparent view of how high-level instructions are grounded into action sequences.

\begin{figure*}[h]
    \centering
    \includegraphics[width=0.9\linewidth]{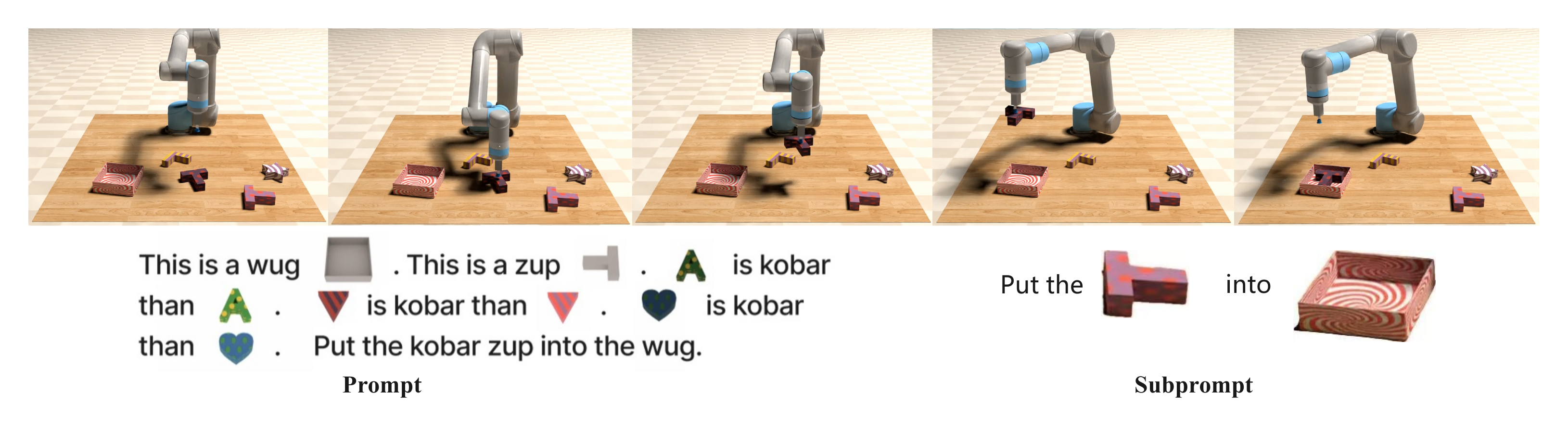}\\[0.5em]
    \includegraphics[width=0.9\linewidth]{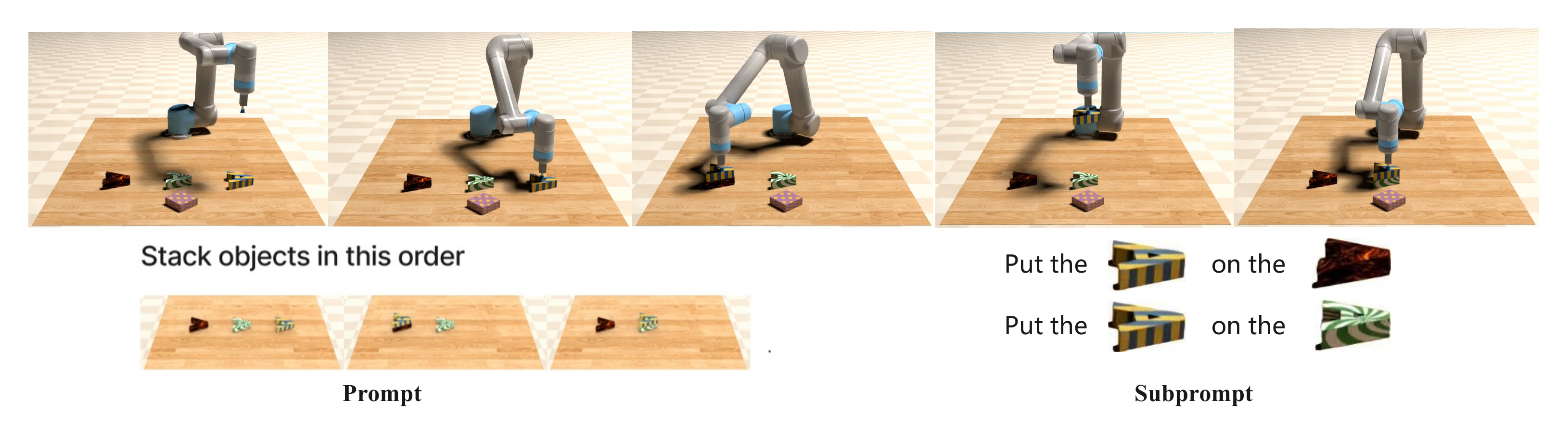}\\[0.5em]
    \includegraphics[width=0.9\linewidth]{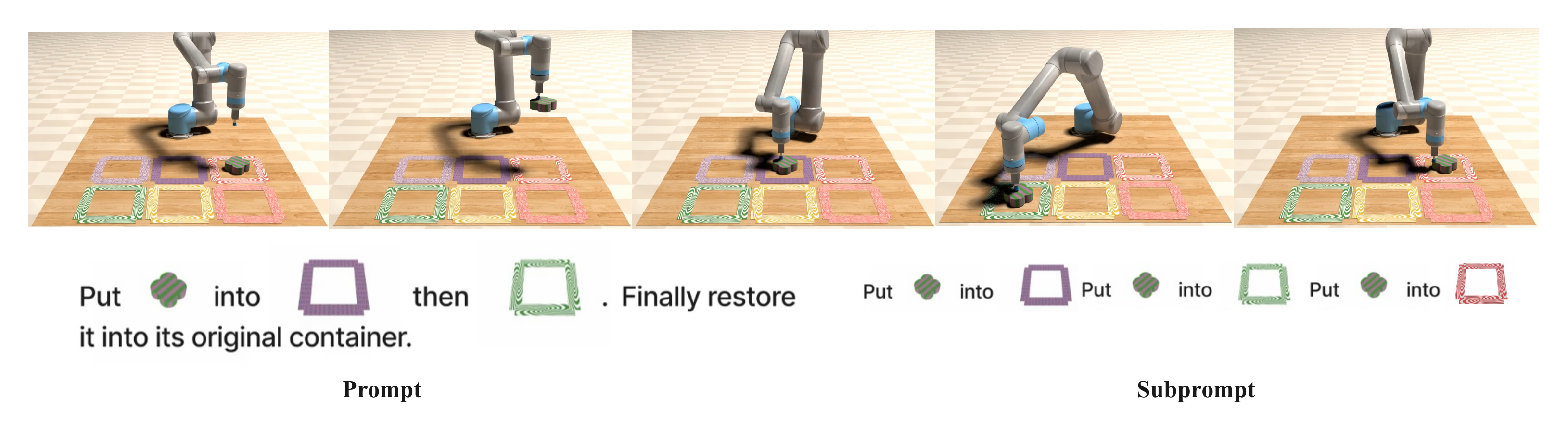}
    \caption{
    Example of VIMA-Bench.
    }
    \label{fig:vima_qual}
\end{figure*}

\begin{figure*}[h]
    \centering
    \includegraphics[width=1\columnwidth]{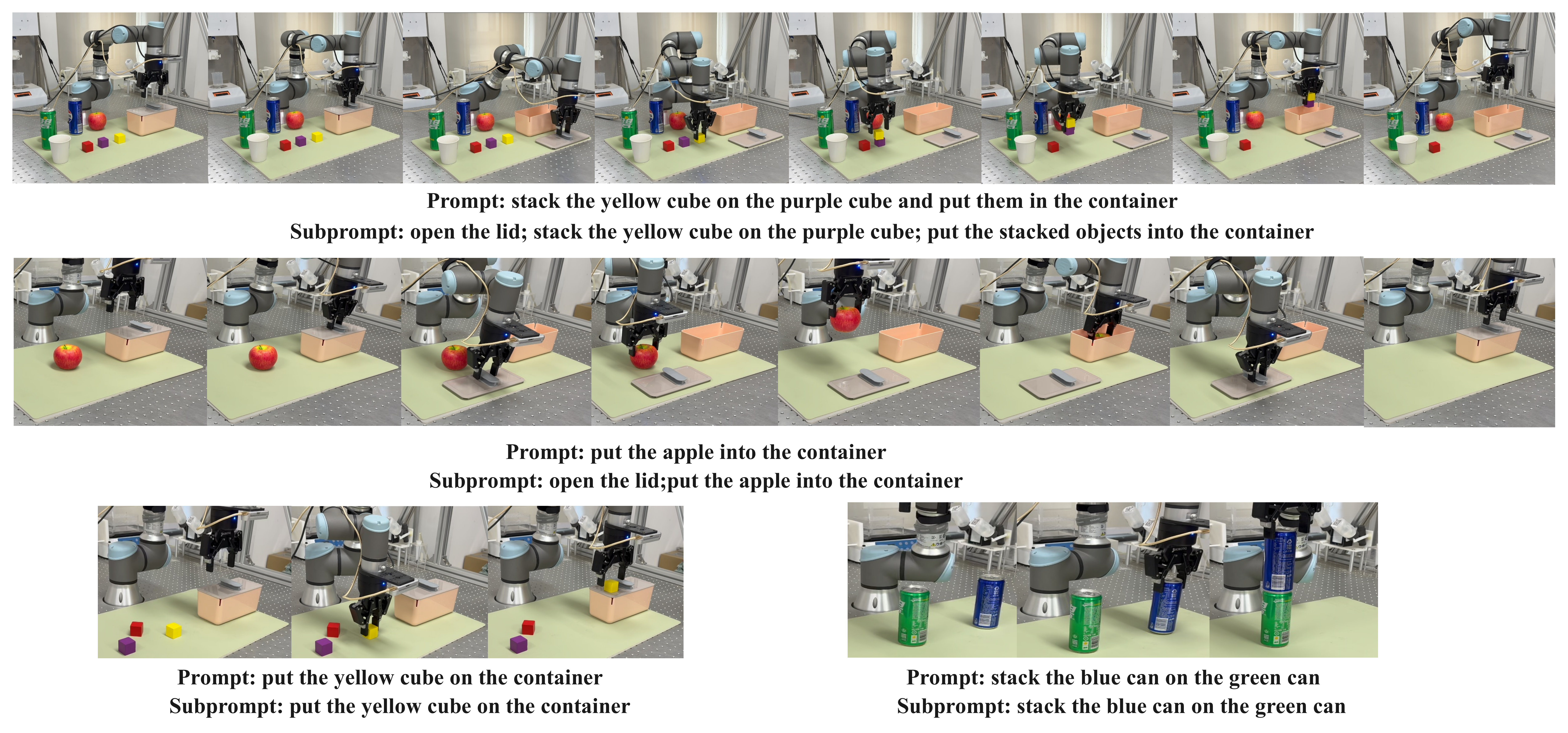}
    \caption{
    Example of the real-world experiment.
    }
    \label{fig:realworld_qual}
\end{figure*}

We include examples from both the VIMA benchmark (Figure \ref{fig:vima_qual}) and real-world experiments (Figure \ref{fig:realworld_qual}).

\begin{figure*}[h]
    \centering
    \includegraphics[width=1\columnwidth]{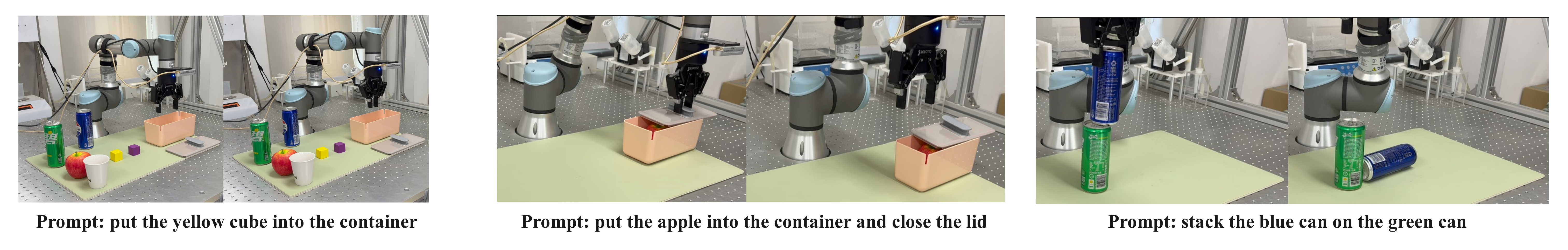}
    \caption{
    Failure case of the real-world experiment.
    }
    \label{fig:fail}
\end{figure*}

We additionally show failure cases from real-world experiments, as shown in Figure~\ref{fig:fail}.
The figure highlights several typical failure modes observed during execution.

The first failure case is an incorrect prerequisite assessment, where the policy attempts to open a container that is already open.
This behavior indicates residual limitations in reliably grounding object states from visual observations under real-world conditions.
The second failure mode arises from spatial misalignment, where the manipulated object is not properly aligned with the target location.
Such errors are often amplified by physical uncertainties, including imperfect grasping, slippage, and calibration noise, and can lead to cascading failures in subsequent steps.
Together, these cases illustrate that while our method improves high-level task reasoning and execution structure, robustness to fine-grained state perception and precise spatial alignment remains a key challenge in real-world robotic manipulation.

\end{document}